\documentclass[10pt,twocolumn,letterpaper]{article}

\usepackage[pagenumbers]{cvpr} 

\usepackage[accsupp]{axessibility}  
\usepackage{multirow} 
\usepackage{multicol} 
\usepackage{graphicx} 
\usepackage{caption}
\usepackage{subcaption}

\usepackage{colortbl}  
\usepackage{xcolor}
\usepackage{pifont}
\usepackage{array}   
\usepackage{amsmath}
\usepackage{float}
\newcommand{\True}{\textcolor{green}{\ding{51}}}
\newcommand{\False}{\textcolor{red}{\ding{55}}}
\definecolor{Gray}{gray}{0.94}

\usepackage[frozencache,cachedir=.]{minted}

\definecolor{cvprblue}{rgb}{0.21,0.49,0.74}
\usepackage[pagebackref,breaklinks,colorlinks,allcolors=cvprblue]{hyperref}

\def\paperID{3958} 
\def\confName{CVPR}
\def\confYear{2025}

\def\modelname{Holmes-VAU}
\title{ \textcolor{black}{\modelname: Towards Long-term Video Anomaly Understanding \\ at Any Granularity}}%

\author{
    Huaxin Zhang$^{1}$
    \hspace{0.02cm}
    Xiaohao Xu$^2$
    \hspace{0.02cm}
    Xiang Wang$^1$
    \hspace{0.02cm}
    Jialong Zuo$^1$
    \hspace{0.02cm}
    Xiaonan Huang$^2$
    \hspace{0.02cm}
    Changxin Gao$^1$ \\
    \hspace{0.02cm}
    Shanjun Zhang$^3$
    \hspace{0.02cm}
    Li Yu$^1$
    \hspace{0.02cm}
    Nong Sang$^{1}\thanks{Corresponding author}$\\
    $^1$Key Laboratory of Image Processing and Intelligent Control, \\
    School of Artificial Intelligence and Automation, Huazhong University of Science and Technology \\
    $^2$University of Michigan, Ann Arbor
    \hspace{0.5cm}
    $^3$Kanagawa University\\
    {\tt\scriptsize \{zhanghuaxin,wxiang,cgao,hustlyu,nsang\}@hust.edu.cn, \{xiaohaox,xiaonanh\}@umich.edu, \{chiyoz01\}@kanagawa-u.ac.jp} \\
}

\begin{document}
\maketitle
\begin{abstract}

How can we enable models to comprehend video anomalies occurring over varying temporal scales and contexts?
Traditional Video Anomaly Understanding (VAU) methods focus on frame-level anomaly prediction, often missing the interpretability of complex and diverse real-world anomalies. Recent multimodal approaches leverage visual and textual data but lack hierarchical annotations that capture both short-term and long-term anomalies.
To address this challenge, we introduce HIVAU-70k, a large-scale benchmark for hierarchical video anomaly understanding across any granularity. We develop a semi-automated annotation engine that efficiently scales high-quality annotations by combining manual video segmentation with recursive free-text annotation using large language models (LLMs). This results in over 70,000 multi-granular annotations organized at clip-level, event-level, and video-level segments.
For efficient anomaly detection in long videos, we propose the Anomaly-focused Temporal Sampler (ATS). ATS integrates an anomaly scorer with a density-aware sampler to adaptively select frames based on anomaly scores, ensuring that the multimodal LLM concentrates on anomaly-rich regions, which significantly enhances both efficiency and accuracy.
Extensive experiments demonstrate that our hierarchical instruction data markedly improves anomaly comprehension. The integrated ATS and visual-language model outperform traditional methods in processing long videos.
Our benchmark and model are publicly available at \url{https://github.com/pipixin321/HolmesVAU}.
\end{abstract}

\vspace{-4mm}
\section{Introduction}
\label{sec:introduction}

Video Anomaly Understanding (VAU) is crucial for applications such as video surveillance~\cite{ucf}, violent content analysis~\cite{xdviolence}, and autonomous driving~\cite{yao2022dota}.
Detecting deviations from normal patterns aids in hazard prevention and real-time decision-making.
Traditional methods~\cite{ConvAE, rtfm, URDMU} mainly focus on frame-level predefined closed-set anomaly prediction, assigning an anomaly score to each frame. However, these approaches often fail to describe and understand complex anomalies in the real world.

To address this gap, open-world anomaly understanding~\cite{wu2024open} embraces the diversity and unpredictability of real-world anomalies. Recent work integrates multimodal approaches, combining visual data with textual descriptions~\cite{wu2024vadclip, pu2024learning, yang2024text, yuan2024towards}, while advances in multimodal visual-language models (VLMs)~\cite{liu2024visual, liu2023improved, li2023videochat, zhang2023video, chen2024far} have enabled more nuanced understanding through anomaly-related instruction tuning and text generation~\cite{du2024uncovering, lv2024video, zanella2024harnessing, tang2024hawk}.

Despite these advancements, \textbf{a significant gap remains in models' ability to comprehend anomalies across multiple temporal scales.} For instance, while anomalies such as explosions or fights may be captured in a single frame, more complex events like theft or arson require understanding long-term contextual patterns. Existing VAU datasets~\cite{xdviolence,ucf} typically provide annotations at a single level of granularity, limiting models to understanding either immediate perceptual anomalies or those requiring extended contextual reasoning.
The lack of datasets with hierarchical annotations—encompassing both short-term and long-term anomalies—hinders models' capacity to reason about anomalies with diverse temporal characteristics.
Moreover, constructing datasets that encapsulate this hierarchical complexity poses significant challenges in scalability and annotation quality.

To address these issues, we develop a semi-automated annotation engine that efficiently scales high-quality annotation by combining manual video segmentation with recursive free-text annotation using large language models (LLMs). The process involves three key stages: \textbf{1}) \textit{hierarchical video decoupling}, where we manually identify anomaly events and segment them into shorter clips; \textbf{2}) \textit{hierarchical free-text annotation}, where captions for each clip are generated through human effort or video captioning models, then summarized at the event and video levels via LLMs; and \textbf{3}) \textit{hierarchical instruction construction}, where the textual data is transformed into question-answer instruction prompts by combining captions and summaries with designed prompts, creating a dataset with rich annotations for training and evaluating models.

Utilizing the annotation engine, we introduce \textbf{\textit{HIVAU-70k}}, a large-scale video anomaly understanding benchmark with hierarchical instructions. Our dataset comprises over 70,000 multi-granular instruction data organized across clip-level, event-level, and video-level segments as shown in ~\cref{fig:motivation}. This hierarchical structure empowers models to detect immediate anomalies, \textit{e.g.}, sudden explosions or fighting, as well as complex events that require an understanding of long-term context, like theft or arson. 
By annotating at multiple temporal levels, HIVAU-70k provides diverse anomalies in open-world scenarios.

Towards long-term VAU, efficiency  remains a critical challenge. Previous methods~\cite{zanella2024harnessing,du2024uncovering,tang2024hawk} often rely on \textit{uniform frame sampling}, which can either miss crucial anomaly frames or incur large computational costs~\cite{lin2023video, zanella2024harnessing, tang2024hawk}. To address this, we propose the \textbf{\textit{Holmes-VAU}} method, which combines the proposed Anomaly-focused Temporal Sampler (ATS) with the multimodal visual-language model for efficient long-term video anomaly understanding (See Fig.~\ref{fig:motivation}). The ATS combines a \textit{anomaly scorer} with a \textit{density-aware sampler} that adaptively selects frames by their anomaly scores. This integration ensures that the visual-language model concentrates on anomaly-rich regions, enhancing both efficiency and accuracy.

Our contributions are threefold: \textbf{1)} We introduce \textit{\textbf{HIVAU-70k}}, a large-scale, multi-granular benchmark for hierarchical video anomaly understanding.
\textbf{2)} We propose the \textit{\textbf{Holmes-VAU}} method, which combines the proposed Anomaly-focused Temporal Sampler (ATS) to boost the efficiency and accuracy of inference on long-term videos.
\textbf{3)} We conduct extensive experiments demonstrating the effectiveness of hierarchical instruction data in enhancing anomaly comprehension and validate the performance gains provided by the integrated ATS and visual-language model in processing long videos.

\section{Related Works}

\noindent\textbf{Video Anomaly Detection.}
This task aims to temporally detect abnormal frames in a long untrimmed video~\citep{adam2008robust,mehran2009abnormal,li2013anomaly,Luetal,GODs,ConvAE}.
Early attempts are based on hand-crafted features~\citep{adam2008robust,kim2009observe,zhao2011online,mehran2009abnormal,Luetal,li2013anomaly}.
Recently, deep learning approaches~\citep{ConvAE,yang2023video,UMIL} have become dominant, 
broadly classified into unsupervised, weakly-supervised, and fully-supervised methods. Unsupervised methods train only on normal videos to learn normal patterns and are often designed as reconstruction-based~\citep{ConvAE,xu2017detecting,gong2019memorizing,yang2023video}, prediction-based~\citep{framepred}, or a combination~\citep{liu2021hybrid}. Some methods~\citep{zaheer2022generative,thakare2023dyannet,tur2023exploring} also explore a fully unsupervised setting, including both normal and abnormal videos in training set without real labels.
Weakly-supervised methods~\citep{ucf,GCN,mist,xdviolence,rtfm,MSL,S3R,URDMU,UMIL,zhang2024glancevad} use both normal and abnormal videos with video-level annotations. Fully-supervised methods~\citep{liu2019exploring,landi2019anomaly} are less studied due to the high cost of precise frame-level annotations.

\vspace{1mm}
\noindent\textbf{Multi-modal Video Anomaly Understanding.}
Large-scale visual-language pre-trained models such as CLIP~\citep{radford2021learning} serve as a bridge between visual and textual modalities.
Some recent works~\citep{pu2024learning,cliptsa,wu2024vadclip,yang2024text} in the realm of video anomaly detection
have leveraged textual information as prompts to enhance the model's anomaly representation.
Based on this, ~\cite{wu2024open} firstly proposed the open vocabulary VAD task,  ~\cite{yuan2024towards} introduced a multimodal video anomaly dataset composed of dense clip captions.
Furthermore, ~\cite{zanella2024harnessing} extracted captions from video frames and designed prompts for LLMs to provide anomaly scores, ~\cite{du2024uncovering} and ~\cite{tang2024hawk} construct diverse and interactive instruction data at the video level.
However, these datasets consider only a single temporal level of anomaly understanding data construction, \ie clip-level~\cite{yuan2024towards} or video-level~\cite{du2024uncovering,tang2024hawk}.
Unlike these methods, we focus on building large-scale \textit{hierarchical} video anomaly understanding data for multimodal instruction tuning.

\vspace{1mm}
\noindent\textbf{Hierarchical Video Understanding.}
Video understanding is a challenging task due to its temporal-scale diversity. To better comprehend videos, many previous works have focused on both datasets and models in hierarchical video understanding. For example, ~\cite{li2018resound,shao2020finegym,liu2022fineaction,ashutosh2023hiervl,wang2022review,lu2023review} proposed fine-grained action recognition and localization datasets, ~\cite{islam2024video} provided free-form hierarchical captions for hour-long videos at multiple temporal scales, and ~\cite{kahatapitiya2021coarse,qing2022learning,dang2024adaptive,xu2022towards,hrpro,TGC-Net} trained models on hierarchical levels to obtain better video feature representation. Recently, to assess the capability of video vision-language models (VLMs) in handling challenges in real-world scenarios, several video benchmarks~\cite{fu2024video,fang2024mmbench} have been built, incorporating data at multiple temporal scales for evaluation.
Unlike these works, we dive into the field of Video Anomaly Understanding, thus filling the gap in multi-scale annotations in this area.

\section{HIVAU-70k Benchmark}
We first define the video anomaly understanding task in Sec.\ref{sec:task}.
Then, the construction process of the HIVAU-70k benchmark will be elaborated in Sec.\ref{sec:data_engine}.
Finally, we will present HIVAU-70k's statistical information in Sec.~\ref{sec:data_statistic}.

\subsection{{Task Description}}
\label{sec:task}
This work focuses on video anomaly understanding, involving temporal anomaly detection and anomaly explainability.
Temporal anomaly detection aims to predict an anomaly score for each frame in a video $\mathcal{V}$, represented as $S \in \mathbb{R}^{T}$, where $T$ is the total number of frames. Building on this, we explore the model's ability to generate explanatory text outputs related to anomalies based on video input and user queries. Specifically, given a video $\mathcal{V}$ and a text query $\mathcal{Q}$, the model generates an anomaly-related response $\mathcal{A}$. We consider two key abilities: \textbf{1}) \textbf{visual perception}, which involves recognizing main entities in the video, and \textbf{2}) \textbf{anomaly reasoning}, which encompasses the model's judgment and analysis of the anomaly content.

\subsection{LLM-Empowered Data Engine}
\label{sec:data_engine}
\begin{figure}[t!]
\centering\setlength{\abovecaptionskip}{0.1cm}
\includegraphics[width=0.48\textwidth]{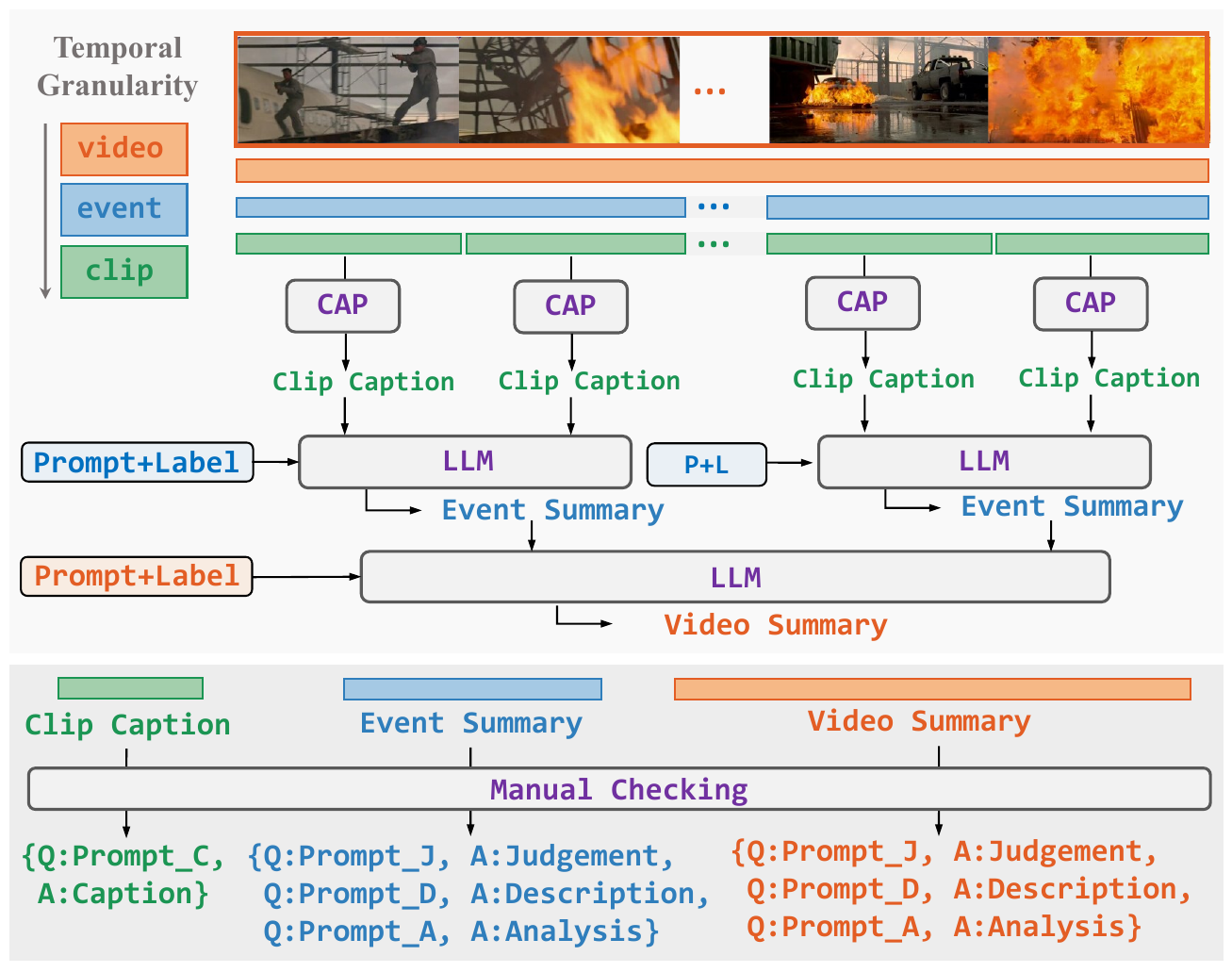}
\caption{\textbf{Data Engine}. We present a structured workflow for generating hierarchical annotations across video, event, and clip levels. Clips are first captioned, then processed through a large language model (LLM) with prompts for event summarization. The outputs include clip captions, event summaries, and video summaries, followed by manual checking and refinement. This multi-step approach enriches the dataset with detailed judgments, descriptions, and analyses of anomalies, enabling robust contextual understanding at varying granularities.}
\vspace{-2mm}
\label{fig:data_engine}
\end{figure}

\begin{figure*}[ht!]
\centering\setlength{\abovecaptionskip}{0.1cm}
\includegraphics[width=\textwidth]{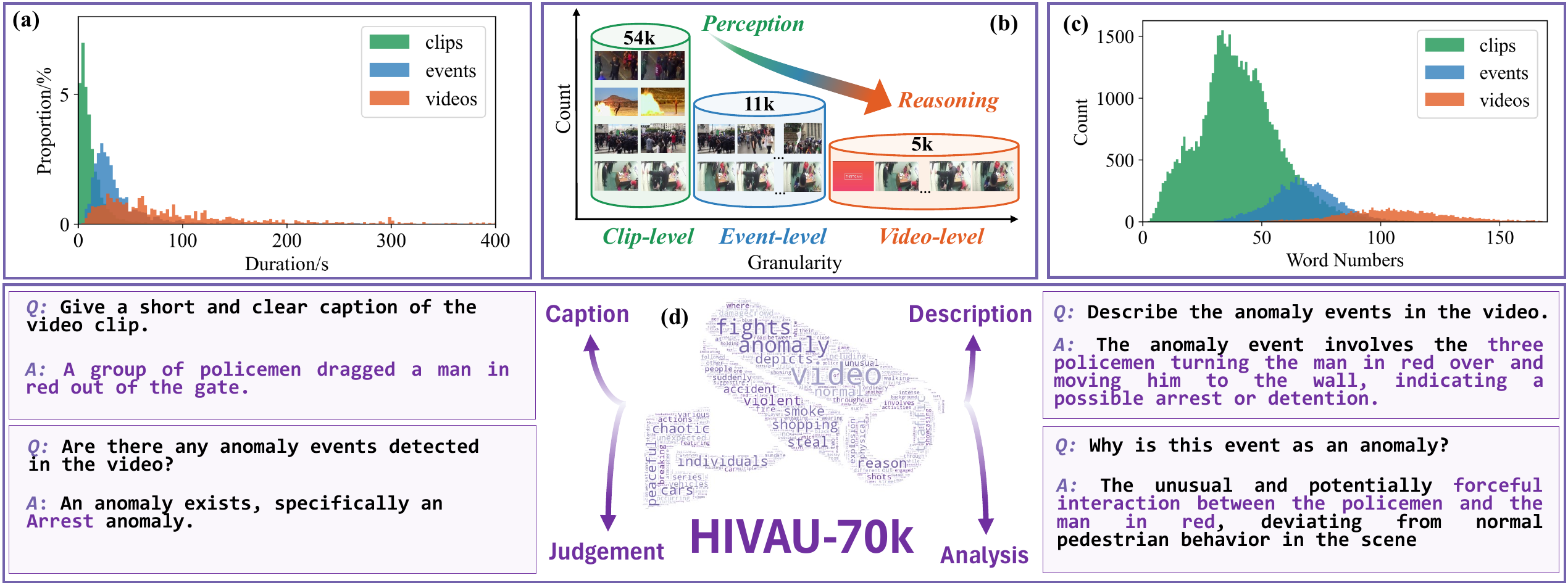}
\caption{\textbf{HIVAU-70k} \textbf{dataset}.  (\textbf{a}) Duration distributions for clips, events, and full videos, showing dominance of short clips. (\textbf{b}) Hierarchical data organization from clip-level to video-level, enabling perception-to-reasoning insights. (\textbf{c}) Word count variations across annotation levels, with more detailed descriptions at the video level. (\textbf{d}) Sample annotations capturing captioning, judgment, description, and anomaly analysis, highlighting nuanced understanding of anomaly events in complex scenes.}
\vspace{-2mm}
\label{fig:data_stastic}
\end{figure*}

As shown in Fig.\ref{fig:data_engine}, we develop a semi-automated annotation engine that efficiently scales high-quality annotations, which consists of three main steps: 1) hierarchical video decoupling, 2) hierarchical free-text annotation, and 3) hierarchical instruction construction. 

\noindent\textbf{Hierarchical Video Decoupling.}
Our video sources include the training set of the UCF-Crime~\cite{ucf} dataset and the XD-Violence~\cite{xdviolence} dataset, which contains videos of varying durations and diverse real-world anomalies.
For abnormal videos, we first manually obtain the temporal boundaries of each anomaly event in the video. Non-continuous anomalous states are considered separate events.
Then, we divide each event into clips of random lengths.
For normal videos, we apply random sampling to obtain corresponding segments of varying granularities.
Ultimately, we obtained 5,443 videos, 11,076 events, and 55,806 clips. This process took 5 annotators approximately 20 hours to complete. More details can be found in the appendix.

\noindent\textbf{Hierarchical Free-text Annotation.}
To fully extract semantic information from the clip-level videos, we utilize a powerful off-the-shell video perception model LLaVA-Next-Video~\cite{liu2024llavanext} to generate detailed captions for each clip. We also include the UCA dataset ~\citep{yuan2023surveillance}, which provides manually annotated captions for video clips in the UCF-Crime ~\citep{ucf} dataset.
Then, we use an LLM~\cite{llama3modelcard} to consolidate all clip captions within an event, generating an event-level video summary. Specifically, we design prompts\footnote{For detailed prompts, please refer to the appendix.} to guide the LLM to produce three parts of content for each anomalous event summary: \textbf{1}) \textit{Judgment}: A determination of whether an anomaly exists and its specific category, \textbf{2}) \textit{Description}: A detailed description of the anomalous or normal event, \textbf{3}) \textit{Analysis}: The reasoning behind the anomaly judgment, including causal analysis.
To guide the LLM in generating reliable responses, we also inject the event's category label (\eg, "Shooting", "Explosion") into the LLM prompt.
We consolidate all event-level summaries to obtain the video-level summary. This results in free-text annotations across multiple temporal scales, including short-term visual perception (clip-level) and long-term anomaly reasoning (event-level, video-level).

\noindent\textbf{Hierarchical Instruction Data Construction.}
The ability of VLMs to follow user instructions and generate responses is achieved through instruction-tuning~\cite{liu2024visual,liu2023improved,li2023videochat,lin2023video}. The training data format typically consists of the following:

`\{$\mathcal{Q}$:\textit{user instruction}, $\mathcal{A}$:\textit{model response.}\}'

To build instruction-tuning data for VLMs in the domain of Video Anomaly Understanding,
we matched free-text annotations with pre-designed anomaly-related user instructions.
Specifically, for clip-level segments, we only construct instructions related to captions, as it is challenging to obtain anomaly-related analysis for short videos. For event-level and video-level segments, we construct instruction data from the perspectives of Judgment, Description, and Analysis. Typical examples are shown in the Fig.~\ref{fig:data_stastic} (d).

\noindent\textbf{Manual Checking for Data Quality Control.}
To ensure dataset quality, we implemented several human inspection and curation strategies. First, we labeled the temporal boundaries of abnormal event segments and reviewed them at the second level. Next, anomaly labels were incorporated during summary generation using LLMs, directing focus on relevant entities. Finally, manual reviews were performed to  correct low-quality instruction data.

\subsection{Data Statistic of HIVAU-70k}
\label{sec:data_statistic}
Utilizing the proposed annotation engine, we introduce HIVAU-70k, as shown in Fig.~\ref{fig:data_stastic}, a large-scale benchmark designed for hierarchical instruction-based video anomaly understanding.
As shown in Fig.~\ref{fig:data_stastic}(b), HIVAU-70k contains over 70,000 multi-granular annotations organized at clip-level, event-level, and video-level segments, achieving a progression from perception to reasoning.
As shown in Fig.~\ref{fig:data_stastic}(a) and (c), the durations of segments and the word numbers of text annotations at different granularities exhibit significant distributional differences.
As shown in Fig.~\ref{fig:data_stastic}(d), HIVAU-70k's instruction data covers \textit{Caption}, \textit{Judgment}, \textit{Description}, and \textit{Analysis} for real-world anomalies, which guide the model to develop both short-term and long-term video anomaly understanding capabilities.

\begin{figure}[t!]
\centering\setlength{\abovecaptionskip}{0.1cm}
\includegraphics[width=0.48\textwidth]{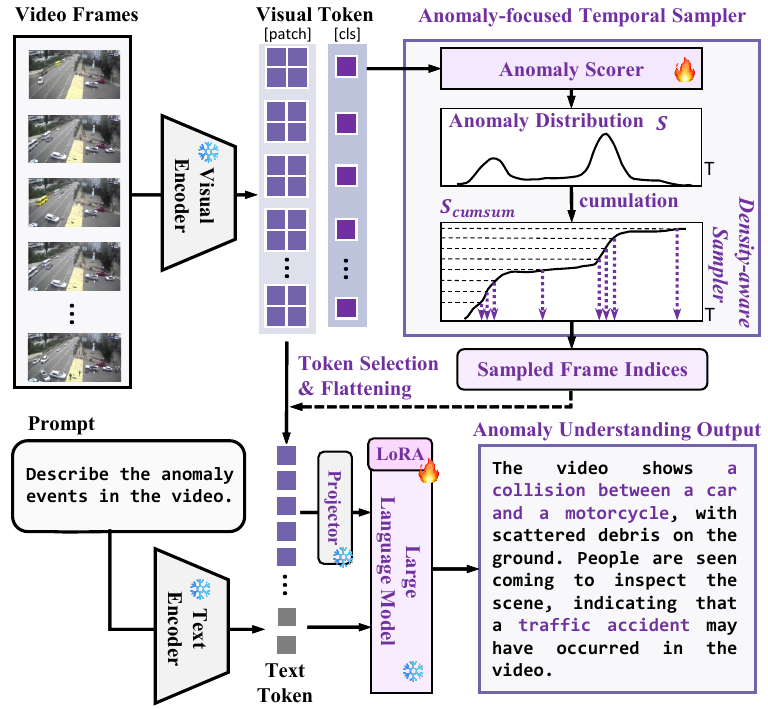}
\caption{\textbf{\modelname:
a multi-modal-LLM-based video anomaly detection framework with adaptive anomaly focus.} 
}
\vspace{-2mm}
\label{fig:framework}
\end{figure}

\section{Method: Holmes-VAU}
Long-term video anomaly understanding with LLMs/VLMs has traditionally been hindered by frame redundancy, complicating accurate anomaly detection. Previous VAU approaches struggle with focus: methods like dense window sampling \cite{zanella2024harnessing} add redundancy, and uniform frame sampling \cite{du2024uncovering, tang2024hawk} often misses key anomalies, limiting application to short videos. We introduce the Anomaly-focused Temporal Sampler (ATS) to address this, integrating it into the VLM, and fine-tuning it via instruction on HIVAU-70k to form our \modelname ~model. 

\subsection{Pipeline Overview}
The overall pipeline of our \modelname ~model is shown in Fig.~\ref{fig:framework}.
Video frames are processed by a visual encoder, creating visual tokens. These tokens are analyzed by an Anomaly-focused Temporal Sampler using an anomaly scorer and cumulative sum ($S_{cumsum}$) to select keyframes. A text encoder processes a prompt (e.g., 'Describe the abnormal events in the video'). Visual and textual representations are combined in a pre-trained language model, fine-tuned with LoRA, to generate a description of detected anomalies, such as a 'collision between a car and a motorcycle' and 'traffic accident' indicators.

\subsection{Model Architecture}
\label{sec:model_architecture}

\noindent\textbf{Visual and Text embedding.}
We utilize the frozen visual encoder in InternVL2~\citep{chen2024far}, which inherits the ViT structure from CLIP~\citep{radford2021learning}, we refer to it as $\phi_v$.
Following previous VAU works~\cite{S3R,URDMU,zanella2024harnessing,wu2024vadclip}, we sample dense video frames at an interval of 16 frames from the input video. Each video frame is then processed through the visual encoder to obtain the corresponding visual tokens. Given the input video frame sequence ${\mathcal{V} \in {\mathbb R}^{T \times H \times W \times C}}$, the output features of $i$-th frame can be denotes as:
\begin{equation}
    V_i=\{v_i^{cls}, v_i^1, v_i^2,...,v_i^{N_p}\} = \phi_v(\mathcal{V}_i)
\end{equation}
where $v_i^{cls}$ indicates the class token,  $v_i^{j}$ ($j \in \{1,2,...,N_p\}$) denotes the patch tokens, and $N_p$ reperesents the number of patches.
The text encoder $\phi_t$ is also initialized from ~\citep{chen2024far}, which includes a tokenizer and an embedding layer. The prompt text $\mathcal{Q}$ is converted into text tokens through the text encoder: $X_q=\phi_t(\mathcal{Q})$.

\noindent\textbf{Anomaly-focused Temporal Sampler (ATS).}
ATS consists of two components: the \textit{anomaly scorer} and the \textit{density-aware sampler}.
The \textit{anomaly scorer} $\phi_{s}$ is a feature-based VAD network which estimates the anomaly score for each frame. We follow the network architecture in ~\cite{URDMU} due to its simplicity and good performance.
Given the class token of the video frames $\{v_1^{cls},v_2^{cls},...,v_T^{cls}\}$, the anomaly scores can be obtained: $s_i = \phi_s(v_i^{cls})$, where $s_i$ denotes the anomaly score of the $i$-th frame.

Anomalous frames typically contain more information and exhibit greater variation than normal frames~\cite{rtfm}. This observation motivates us to sample more frames in regions with higher anomaly scores while reducing the sampling in areas with lower anomaly scores.
As shown in Fig.~\ref{fig:framework}, to achieve non-uniform sampling, we propose \textit{density-aware sampler} to selectively choose $N$ frames from a total of $T$ input frames. Specifically, we treat the anomaly scores $S \in \mathbb{R}^{T}$ as a probability mass function and first accumulate them along the temporal dimension to obtain the cumulative distribution function, denoted as $S_{cumsum}$:
\begin{equation}
S_{cumsum}(t) = \sum_{i=1}^t (s_i + \tau)
\label{eq:cumsum}
\end{equation}
We uniformly sample $N$ points along the cumulative axis, then map these points to the cumulative distribution $S_{cumsum}$, the corresponding $N$ timestamps on the time axis are mapped to the closest frame index and finally form the sampled frame indices, denoted as $\mathcal{G}$. $\tau$ is used to control the uniformity of the sampling.

\noindent\textbf{Projector and LLM.}
We select the tokens corresponding to the sampled frame, \ie, $\mathcal{G}$, as the visual embedding. A projector $\phi_p$ is then used to map the visual embedding to the language feature space. Finally, we concatenate these embeddings with the text embeddings, input them into the pre-trained large language model, and compute the probability of the target answers $X_a$.
To obtain an initial visual-language alignment space, we initialize the projector and LLM parameters from ~\cite{chen2024far}, with the parameters kept frozen.
The above process can be expressed as follows:
\begin{equation}
X_{ins} = \text{cat}[\phi_p(\text{cat}[{V_i}]), X_q] \qquad i \in \mathcal{G}
\end{equation}
\begin{equation}
p(X_a|X_{ins}) = \prod_{i=1}^{L}p_{\theta}|(x_i|X_{ins,<i},X_{a,<i})
\end{equation}
where cat[$\cdot$] represents the concatenation operation, $\theta$ is the trainable parameters, $X_{ins,<i},X_{a,<i}$ are the instruction and answer tokens in all turns before the current prediction token $x_i$, $L$ is the length of sequence, respectively.

\subsection{Training and Testing}
\label{sec:train_test}

\noindent\textbf{Training.}
We train the model following two steps.
In the first step, we use the video data and annotated frame-level label ($\hat{y}\in  {\mathbb R}^{T}$) from HIVAU-70k to train the \textit{anomaly scorer}, which provides more accurate anomaly supervision compared to previous unsupervised and weakly-supervised methods~\cite{ConvAE,ucf,S3R,URDMU}.
\begin{equation}
{\mathcal L}_{AS} = - \sum_{i=1}^{T}( s_ilog(\hat{y}_i) + (1-s_i)log(1-\hat{y}_i))
\end{equation}
In the second step, we keep the anomaly scorer fixed, and use all the instruction data from HIVAU-70k to train the model.
To achieve more efficient fine-tuning without disrupting the original capabilities of the LLM, we employ LoRA~\cite{hu2021lora} for fine-tuning, optimizing the cross entropy loss between the predicted and the ground truth tokens.

\noindent\textbf{Testing.}
During testing, the user inputs a video and text prompts; the model will generate the corresponding text response following the user's instruction.

\section{Experiments}
\label{sec:exp}
\subsection{Experiment Setup}
\noindent\textbf{Dataset.}
Our HIVAU-70k is built upon two large-scale real-world datasets, \ie, UCF-Crime~\cite{ucf} and XD-Violence~\cite{xdviolence}, they provide a diverse range of videos with anomalous events.
UCF-Crime~\citep{ucf} comprises 1,900 untrimmed videos totaling 128 hours from outdoor and indoor surveillance cameras. It encompasses 13 classes of real-world anomalies, including \textit{Abuse}, \textit{Explosion}, \textit{Fighting}, and \textit{Shooting}.
XD-Violence~\citep{xdviolence} is the largest VAD benchmark, comprising 4,754 videos totaling 217 hours sourced from surveillance, movies, car cameras, and games. It encompasses 6 anomaly classes: \textit{Abuse}, \textit{Car Accidents}, \textit{Explosions}, \textit{Fighting}, \textit{Riots}, and \textit{Shooting}.

\noindent\textbf{Metric.} We assess the anomaly understanding ability from two aspects: \textbf{\textit{anomaly} }\textbf{\textit{detection}} and \textbf{\textit{reasoning}}.
\textbf{1})  For \textbf{\textit{anomaly detection}}, we use the anomaly scores output by the Anomaly Scorer as the prediction and perform the evaluation. Following ~\cite{ConvAE,ucf,URDMU,zanella2024harnessing}, we use AUC and AP to quantify detection performance, which is evaluated only on the video level.
\textbf{2})  For \textbf{\textit{anomaly reasoning}}, we annotate instruction data from the UCF-Crime and XD-Violence test sets, which have been carefully reviewed and filtered by annotators. We finally collected 3,300 test samples at multiple granularities.
The test set contains 2200/732/398 samples at clip/event/video levels.
We calculate metrics including BLEU~\cite{papineni2002bleu}, CIDEr~\cite{vedantam2015cider}, METEOR~\cite{banerjee2005meteor} and ROUGE~\cite{lin2004rouge} to measure the quality of the reasoning text output by the model, comparing with the annotated ground truth text.

\noindent\textbf{Implementation Details.}
\begin{table}[t]
\centering\setlength{\abovecaptionskip}{0.1cm}
\caption{
\textbf{Comparison of detection performance with state-of-the-art Video Anomaly Detection approaches}.
We include the results of explainable and non-explainable methods.
"$\ast$" represents the result reported in ~\cite{zanella2024harnessing}.}
\label{tab:vad_comparison}
 \centering\setlength{\tabcolsep}{2mm}
\resizebox{0.47\textwidth}{!}{
\begin{tabular}{l|c|c|c}
\hline
\multicolumn{1}{c|}{\multirow{2}{*}{\textbf{Methods}}}  & \multirow{2}{*}{\textbf{Backbone}}  & \multicolumn{1}{c|}{XD-Violence}  & \multicolumn{1}{c}{UCF-Crime}    \\
\cline{3-4} 
\multicolumn{1}{c|}{} & &  \multicolumn{1}{c|}{AP/\%}  & \multicolumn{1}{c}{AUC/\%}  \\ \hline
\multicolumn{4}{c}{Non-explainable VAD} \\
\hline
Conv-AE~\cite{ConvAE} \textcolor{gray}{\scriptsize (CVPR'16)} & -  & 27.25 & 50.60  \\
GODS~\cite{GODs} \textcolor{gray}{\scriptsize (ICCV'19)} & I3D  & N/A & 70.46  \\
GCL~\cite{zaheer2022generative} \textcolor{gray}{\scriptsize (CVPR'22)} & ResNext  & N/A & 71.04   \\
DYANNET~\cite{thakare2023dyannet} \textcolor{gray}{\scriptsize (WACV'23)} & I3D  & N/A  & 84.50  \\
MIST~\cite{mist} \textcolor{gray}{\scriptsize (CVPR'21)} & I3D  & N/A  & 82.30  \\
Wu~\etal~\cite{xdviolence} \textcolor{gray}{\scriptsize (ECCV'20)} & I3D  & 78.64 & 82.44   \\
RTFM~\cite{rtfm} \textcolor{gray}{\scriptsize (ICCV'21)} & I3D  & 77.81 & 84.30   \\
MSL~\cite{MSL} \textcolor{gray}{\scriptsize (AAAI'22)} & I3D  & 78.28  & 85.30   \\
S3R~\cite{S3R} \textcolor{gray}{\scriptsize (ECCV'22)} & I3D  & 80.26  & 85.99 \\
MGFN~\cite{chen2023mgfn} \textcolor{gray}{\scriptsize (AAAI'23)} & I3D & 79.19 & 86.98  \\
UR-DMU~\cite{URDMU} \textcolor{gray}{\scriptsize (AAAI'23)} & I3D  & 81.66 & 86.97   \\ 
CLIP-TSA~\cite{cliptsa} \textcolor{gray}{\scriptsize (ICIP'23)} & ViT  & 82.19  & 87.58  \\
VadCLIP~\cite{wu2024vadclip} \textcolor{gray}{\scriptsize (AAAI'24)} & ViT  & 84.51  & 88.02  \\
Yang~\etal~\cite{yang2024text} \textcolor{gray}{\scriptsize (CVPR'24)} & ViT  & 83.68  & 87.79 \\
Wu~\etal~\cite{wu2024open} \textcolor{gray}{\scriptsize (CVPR'24)} & ViT & 66.53 & 86.40  \\
\hline
\multicolumn{4}{c}{Explainable Multi-modal VAD} \\
\hline
Zero-Shot CLIP~\cite{radford2021learning}$^*$  & ViT  & 17.83  & 53.16  \\
LLAVA-1.5~\cite{liu2023improved}$^*$  & ViT  & 50.26  & 72.84  \\
LAVAD~\cite{zanella2024harnessing} \textcolor{gray}{\scriptsize (CVPR'24)} & ViT  & 62.01 & 80.28  \\
\textbf{\textit{Holmes}-VAU} \textbf{(Ours)}  & ViT  & \textbf{87.68}  & \textbf{88.96} \\
\hline
\end{tabular}
}\vspace{-2mm}
\end{table}
For the proposed Holmes-VAU method,
we initialize the Multimodal LLM with InternVL2-2B~\cite{chen2024far}.
To optimize the Anomaly-focused Temporal Sampler, we adopt the Adam optimizer with a learning rate of 1e-4.
Note that when evaluating detection performance on XD-Violence and UCF-Crime, only videos in the corresponding training sets are used to train our model for fair comparisons.
For instruction tuning, we train with a batch size of 512 for 1 epoch, using the AdamW optimizer with cosine learning rate decay and a warm-up period. The LoRA~\cite{hu2021lora} parameters are set as: $r$=64, $\alpha$=128, and learning rate=4e-5. 
During testing, $\tau$ in Eq.~\ref{eq:cumsum} is set to 0.1.
Experiments are conducted on 2 NVIDIA A100 GPUs.

\subsection{Main Results}

\begin{table*}[ht!]
\centering\setlength{\abovecaptionskip}{0.1cm}
\caption{
\textbf{Comparison of reasoning performance with state-of-the-art Multimodal Large Language Models (MLLMs).}
'BLEU' refers to the cumulative values from BLEU-1 to BLEU-4.
We evaluate the quality of the generated text at different granularities, including clip-level (C), event-level (E), and video-level (V).
}
\label{tab:reason_comparison}
\resizebox{\textwidth}{!}{
\begin{tabular}{l|c|ccc|ccc|ccc|ccc}
\hline
\multirow{2}{*}{\textbf{Method}} & \multirow{2}{*}{\textbf{Params}}  & \multicolumn{3}{c|}{\textbf{BLEU}~\cite{papineni2002bleu}($\uparrow$)} & \multicolumn{3}{c|}{\textbf{CIDEr}~\cite{vedantam2015cider}($\uparrow$)} & \multicolumn{3}{c|}{\textbf{METEOR}~\cite{banerjee2005meteor}($\uparrow$)} & \multicolumn{3}{c}{\textbf{ROUGE}~\cite{lin2004rouge}($\uparrow$)}  \\
\cline{3-14} 
 &  & C & E & V   & C & E & V  & C & E & V & C & E & V \\ \hline
Video-ChatGPT~\cite{Maaz2023VideoChatGPT} & 7B  & 0.152 & 0.068 & 0.066  & 0.033 & 0.011 & 0.013  & 0.102 & 0.069 & 0.044   & 0.153 & 0.048 & 0.079  \\
Video-LLaMA~\cite{zhang2023video} & 7B & 0.151 & 0.079 & 0.104  & 0.024 & 0.014 & 0.017  & 0.112 & 0.076 & 0.057   & 0.156 & 0.067 & 0.090  \\
Video-LLaVA~\cite{lin2023video} & 7B  & 0.164 & 0.046 & 0.055  & 0.032 & 0.009 & 0.013  & 0.097 & 0.022 & 0.014  & 0.132 & 0.023 & 0.045  \\
LLaVA-Next-Video~\cite{zhang2024video}  & 7B  & 0.435 & 0.091 & 0.120  & 0.102 & 0.015 & 0.031  & 0.117 & 0.085 & 0.096   & 0.198 & 0.080 & 0.106  \\
QwenVL2~\cite{Qwen2VL}  & 7B  & 0.312 & 0.082 & 0.155  & 0.044 & 0.020 & 0.044  & 0.133 & 0.092 & 0.112  & 0.163 & 0.081 & 0.137     \\
InternVL2~\cite{chen2024far}  & 8B  & 0.331 & 0.101 & 0.145  & 0.052 & 0.022 & 0.035  & 0.141 & 0.095 & 0.101  & 0.182 & 0.102 & 0.122     \\
\hline
\textbf{Holmes-VAU (Ours)}  & 2B  & \textbf{0.913} & \textbf{0.804} & \textbf{0.566}   & \textbf{0.467} & \textbf{1.519} & \textbf{1.437}  & \textbf{0.190} & \textbf{0.165} & \textbf{0.121}  & \textbf{0.329} & \textbf{0.370} & \textbf{0.355}   \\
\hline
\end{tabular}
}\vspace{-4mm}
\end{table*}
\begin{table}[t]
\centering
\caption{\textbf{Ablation of hierarchical instruction data.} During the instruction tuning phase, we combined training data of different granularities, including clip (C), event (E), and video (V) levels.}
\vspace{-2mm}
\label{tab:ablation_data}
\resizebox{0.5\textwidth}{!}{
\begin{tabular}{ccc|ccc|ccc}
\hline
\multicolumn{3}{c|}{\textbf{Training Data}} & \multicolumn{3}{c|}{\textbf{BLEU}($\uparrow$)} & \multicolumn{3}{c}{\textbf{CIDEr}($\uparrow$)} \\ \cline{1-9} 
C & E & V  & C & E & V  & C & E & V \\ \hline
$\checkmark$& & &\textbf{0.984}  &0.261  &0.351  &0.459  &0.120  &0.106    \\
&$\checkmark$& &0.508  &0.576  &0.292  &0.097  &1.183  &0.872    \\
&&$\checkmark$ &0.280  &0.222  &0.279  &0.039  &0.708  &0.884    \\
$\checkmark$&$\checkmark$ & &0.889  &0.741  &0.349  &0.470  &1.285  &0.889    \\
$\checkmark$&&$\checkmark$ &0.906  &0.341  &0.522  &\textbf{0.472}  &0.962  &1.093    \\
&$\checkmark$&$\checkmark$  &0.394  &0.797  &0.505  &0.081  &1.472  &1.074    \\
$\checkmark$&$\checkmark$&$\checkmark$ &0.913 &\textbf{0.804} &\textbf{0.566} &0.467 &\textbf{1.519} & \textbf{1.437} \\ \hline
\end{tabular}
} \vspace{-2mm}
\end{table}
\noindent\textbf{Anomaly Detection Results.}
We compare our method with state-of-the-art methods, including semi-supervised methods~\citep{ConvAE,GODs}, unsupervised methods~\citep{zaheer2022generative,thakare2023dyannet}, weakly-supervised methods~\citep{rtfm,MSL,S3R,URDMU,cliptsa,wu2024vadclip} and recently training-free method~\cite{zanella2024harnessing}.
We have indicated their backbones and performance on the UCF-Crime and XD-Violence datasets, as shown in Table~\ref{tab:vad_comparison}.
Our method has an AP of 87.68\% on XD-Violence and an AUC of 88.96\% on UCF-Crime, significantly outperforming the prior state-of-the-art methods,
which demonstrates that our method can generate less biased anomaly scores.
It is worth noting that while achieving precise localization of anomalies, Holmes-VAU is also capable of providing explanations and analysis for the detected anomalies by the model, a feature unavailable in existing non-explainable VAD methods.
Although LAVAD~\cite{zanella2024harnessing} has explainability, the training-free large language model lacks an understanding of anomaly knowledge due to the limitation of insufficient supervised data.

\noindent\textbf{Anomaly Reasoning Results.}
We compare the anomaly-related text quality generated by Holmes-VAU with that produced by state-of-the-art general Multimodal Large Language Models (MLLMs), and presented the results at different temporal granularities, including clip-level, event-level, and video-level, as shown in Table.~\ref{tab:reason_comparison}.
Earlier MLLMs such as Video-ChatGPT~\cite{li2023videochat} and Video-LLaMA~\cite{zhang2023video}, struggled with basic visual perception and instruction-following capabilities.
Recent MLLMs~\cite{zhang2024video,cheng2024videollama,chen2024far} trained on larger and higher-quality video instruction data have made significant progress in general video understanding, with noticeable improvements at the clip-level perception task.
However, due to the absence of learning from complex, real-world anomaly data, their reasoning abilities at the event-level and video-level are still lacking.
Our Holmes-VAU, however, shows significant improvements in video understanding across all temporal granularities compared to existing general MLLMs, highlighting the importance of injecting anomaly-related knowledge through instruction tuning on high-quality Video Anomaly Understanding benchmarks.

\subsection{Analytic Results}

\begin{table}[t]
\centering
\caption{
\textbf{Ablation study of sampling methods and the number of sampled frames.}
We compare the proposed Anomaly-focused Temporal Sampler (ATS) with other sampling methods under different frame sampling numbers, including \textit{Uniform} and \textit{Top-K}.
Latency is the time delay in generating the first token.
}
\vspace{-2mm}
\label{tab:ablation_sampler}
\resizebox{0.48\textwidth}{!}{
\begin{tabular}{c|l|c|cc}
\hline
\multirow{2}{*}{\textbf{Frames (N)}} & \multirow{2}{*}{\textbf{Sampler}} & \multirow{2}{*}{\textbf{Latency (ms)}} & \multicolumn{2}{c}{\textbf{Video-level}} \\ \cline{4-5} 
&  &  & BLEU($\uparrow$) & CIDEr($\uparrow$)     \\ \hline
\multirow{3}{*}{8}
& \textit{Top-K} &   & 0.462 & 1.229    \\
& \textit{Uniform}  &244  & 0.491  & 1.276   \\
& \textbf{ATS (Ours)}   &  & \textbf{0.514} & \textbf{1.324}   \\
\hline
\multirow{3}{*}{16}
& \textit{Top-K} &   & 0.476  & 1.302    \\
& \textit{Uniform}  &566   &0.511  &1.345   \\
& \textbf{ATS (Ours)}  &  &\textbf{0.566}  &\textbf{1.437}   \\
\hline
\multirow{3}{*}{32}
& \textit{Top-K} &   & 0.481  & 1.332    \\
& \textit{Uniform}  & 1402   & 0.558 & 1.357  \\
& \textbf{ATS (Ours)}   &  & \textbf{0.576} & \textbf{1.460}   \\
\hline

\end{tabular}
}\vspace{-2mm}
\end{table}
\begin{figure*}[t]
\centering\setlength{\abovecaptionskip}{0.1cm}
\includegraphics[width=\textwidth]{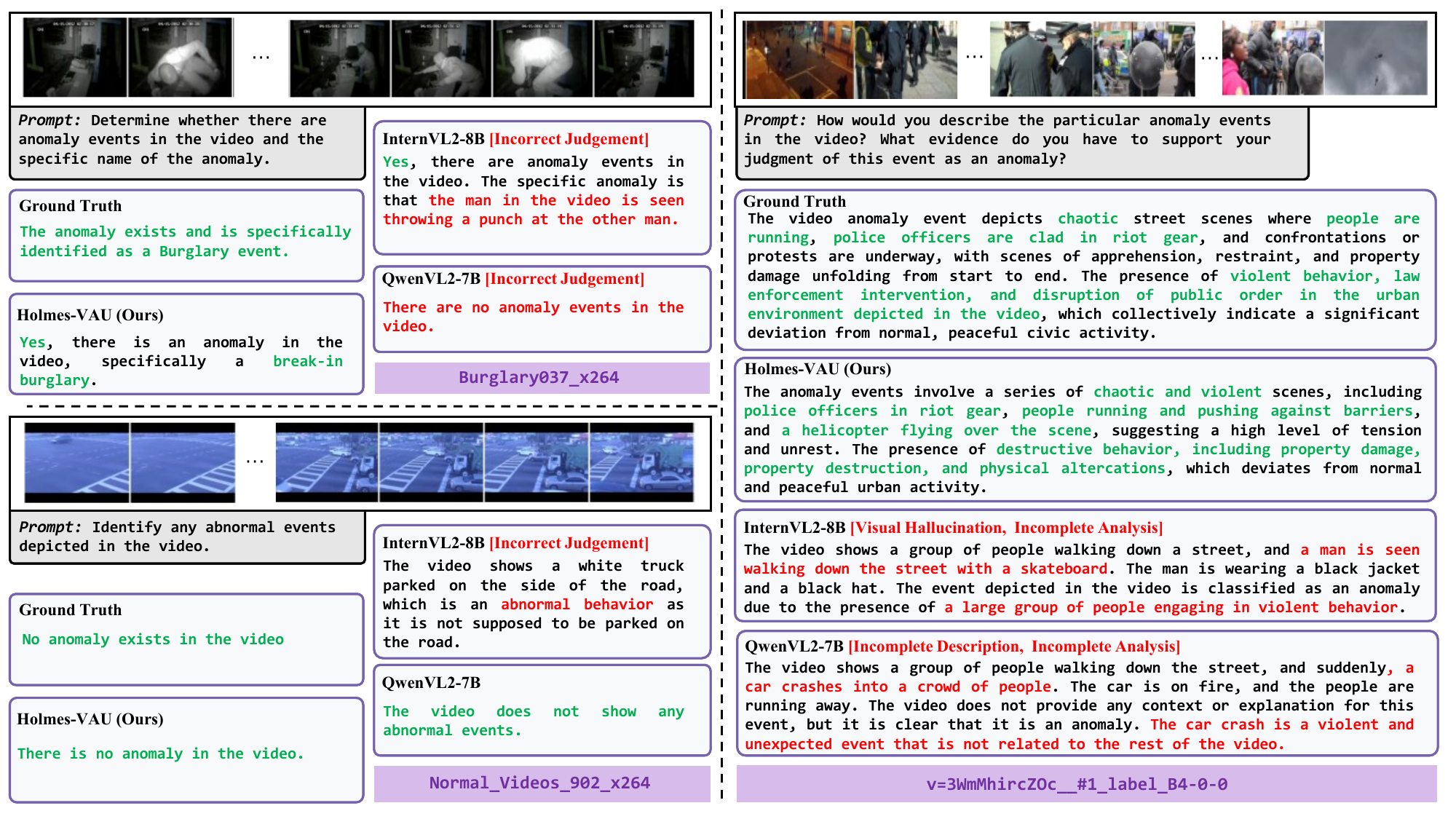}
\caption{\textbf{Qualitative comparison of anomaly understanding explanation.}
Compared with state-of-the-art general MLLMs, \ie, InternVL2~\cite{chen2024far} and QwenVL2~\cite{Qwen2VL}, our proposed Holmes-VAU demonstrates more accurate anomaly judgment, along with more detailed and comprehensive anomaly-related descriptions and analysis.
Correct and wrong explanations are highlighted in \textcolor[rgb]{0,0.6875,0.3125}{green} and \textcolor{red}{red}, respectively.
}
\vspace{-4mm}
\label{fig:qualitative}
\end{figure*}

\noindent\textbf{Influence of Hierarchical Instruction.}
To explore the impact of different granularity video training data on the model's anomaly reasoning ability, we designed various training data combinations during the instruction tuning phase and evaluated the model's performance, as shown in Table~\ref{tab:ablation_data}.
The inclusion of clip-level data primarily enhanced the model's basic visual perception abilities regarding actions and scenes within the video.
Adding event-level data improved the model's ability to judge and understand complete anomaly events.
Furthermore, the involvement of video-level data further enhanced the model's ability to analyze and summarize anomaly-related information across longer-span videos.
The hierarchical instruction data structure facilitated a comprehensive and complementary improvement in the model's anomaly-related perception-to-reasoning capabilities.

\noindent\textbf{Influence of different sampling methods and the number of sampled frames.}
Our ATS (Anomaly-focused Temporal Sampler) is designed to adaptively sample frames input to the LLM based on the frame-level anomaly scores.
To validate its advantages, we compared ATS with other sampling methods at various sample frame counts, including \textit{Uniform} and \textit{Top-K} sampling.
In \textit{Uniform} sampling, N frames are uniformly sampled from all frames, while Top-K sampling selects the frames with the top $N$ highest anomaly scores.
As shown in Table~\ref{tab:ablation_sampler}, ATS consistently outperforms other sampling methods, regardless of the sample count.
We believe that \textit{Uniform} sampling tends to overlook key anomaly frames, though this issue lessens as more frames are sampled. Besides, \textit{Top-K} sampling tends to overly focus on local anomaly frames, missing contextual frame information.
Our proposed ATS mitigates both issues.
To balance inference efficiency and performance, we set $N$=16 as the default sample frame number.

\begin{figure}[t!]
\centering\setlength{\abovecaptionskip}{0.1cm}
\includegraphics[width=0.48\textwidth]{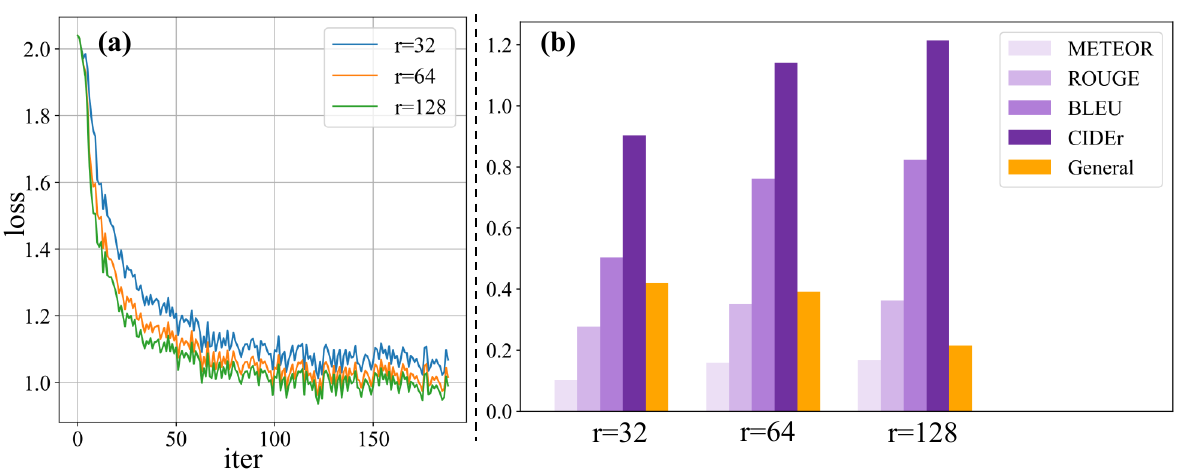}
\caption{
\textbf{Ablation study of trainable parameters.}
\textbf{(a)} Loss curve during instruction-tuning.
\textbf{(b)} We tuned the LoRA~\cite{hu2021lora} parameter $r$ to control trainable parameters, evaluating its impact on VAU capability, and \textit{General} performance on Video-MME~\cite{fu2024video}.
}
\vspace{-3mm}
\label{fig:ablation_lora}
\end{figure}

\noindent\textbf{Instruction Tuning Parameters.}
We conducted an ablation study on the parameter $r$ in LoRA~\cite{hu2021lora} to explore how the trainable parameters affects both the model's VAU performance and its general capability. We use Video-MME~\cite{fu2024video} to evaluate the model's general capability.
The results are shown in Fig.~\ref{fig:ablation_lora}, as $r$ increases, the model gradually adapts to the VAU task. However, when $r$ becomes too large, the model's general capability decreases. To retain the original general video understanding capability of the MLLM, we set $r$=64 as the default value.

\subsection{Qualitative Comparision}
We provide qualitative comparisons between Holmes-VAU and existing MLLMs in Fig.~\ref{fig:qualitative}.
The results demonstrate that Holmes-VAU can accurately identify anomalies in videos and provide accurate and complete explanations, highlight the effectiveness and advantage of Holmes-VAU in perceiving video events and analyzing anomalies.
\section{Conclusion}
In conclusion, this work pushes the boundaries of video anomaly understanding by introducing hierarchical anomaly detection across diverse temporal scales, from momentary clips to extended events. The HIVAU-70k benchmark, with over 70,000 multi-level annotations, addresses a critical gap in the field, enabling comprehensive anomaly analysis in real-world scenarios. Our Anomaly-focused Temporal Sampler (ATS) strategically enhances focus on anomaly-dense segments, optimizing both efficiency and accuracy in long-term anomaly detection. Extensive experiments demonstrate that our hierarchical dataset and ATS-enhanced VLM achieve significant performance gains over conventional methods, proving robust for open-world anomaly understanding. This work sets a new standard for multi-granular anomaly comprehension, paving the way for more fine-grained video anomaly understanding.

\noindent\textbf{Acknowledgement}
This work is supported by the National Natural Science Foundation of China under grants U22B2053 and 623B2039, and in part by the Interdisciplinary Research Program of HUST (2024JCYJ034).

{
    \small
    \bibliographystyle{ieeenat_fullname}
    \bibliography{main}
}

\appendix

\setcounter{table}{0}  
\setcounter{figure}{0}  
\renewcommand{\thetable}{\Alph{table}}
\renewcommand{\thefigure}{\Alph{figure}}

\twocolumn[
{
    \centering
    \Large
    \textbf{\thetitle}\\
    \vspace{0.3em}Supplementary Material \\
    \vspace{36pt}
    }
]

\section{Details of the Data Engine.}
To construct a dataset with hierarchical annotations with both short-term and long-term anomalies, we developed a semi-automated annotation engine that combines manual efforts with the generative capabilities of LLM.
In the main paper, we present the complete annotation workflow. Below, we provide additional details about the data engine.

\subsection{Hierarchical Video Decoupling}
Before annotation, we collected videos from the training sets of the UCF-Crime~\cite{ucf} and XD-Violence~\cite{xdviolence} datasets.
From UCF-Crime, we selected 758 normal videos and 735 anomaly videos, while from XD-Violence, we selected 1,904 normal videos and 2,046 anomaly videos.
The anomaly videos included their original video-level labels, \eg, \textit{Abuse}, \textit{Explosion}.
For the anomaly videos, we organized a team of five annotators to label each anomaly event within the videos. The annotation process took approximately 20 hours to complete.
For the normal videos, we considered all segments to be normal and randomly cropped segments of varying lengths to serve as normal event-level video segments.
These anomaly and normal event-level videos were further divided into shorter clip-level segments. For UCF-Crime, we adopted the clip-level divisions from UCA~\cite{yuan2024towards}. For XD-Violence, we performed uniform division.

\subsection{Hierarchical Free-text Annotation}
\noindent\textbf{Clip Captioning.}
For videos in  UCF-Crime, we fully utilized the manually annotated captions from UCA~\cite{yuan2024towards}. For  videos in XD-Violence, we used LLaVA-Next-Video-7B~\cite{zhang2024video} as our captioner to generate textual descriptions for clip-level videos. The specific prompt is as follows:

\textit{'Please provide a short and brief description of the video clip, focusing on the main subjects and their actions.'}

\noindent\textbf{Event Summary.}
We combined all captions and video-level category labels to generate anomaly-related summaries for each event using an LLM. We selected LLaMA3-70B~\cite{llama3modelcard} as our LLM due to its strong summarization capabilities. The specific prompt is as follows:

\textit{'The dense caption of the video is: \{clip captions\}. There are (is no) abnormal events (\{video-level label\}) in the video. Your response should include the following three parts: 1. Whether the anomaly exists and the specific name of the anomaly. 2. A summary of the anomaly events. 3. Brief explanation of the basis for judging the anomaly.'}

\noindent\textbf{Video Summary.}
Similar to generating event summaries, we generated video-level summaries by analyzing the event-level summaries. The specific prompt is as follows:

\textit{'Below is a summary of all the events in the video: \{event summaries\}. There are (is no) abnormal events (\{video-level label\}) in the video. Your response should include the following three parts: 1. Whether the anomaly exists and the specific name of the anomaly. 2. Detailed description of the video anomaly event from start to end. 3. Brief analysis of the basis for judging the anomaly.'}

\begin{figure*}[h!]
\centering
\includegraphics[width=\textwidth]{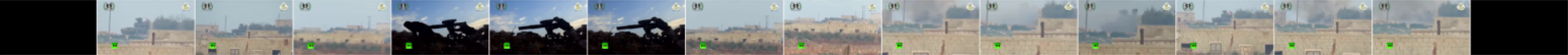}
\vspace{-6mm}
\scriptsize{
\begin{minted}[linenos, frame=lines, breaklines]{json}
{
    "video": "v=2rfyeR-YaJw__#1_label_G-0-0",
    "n_frames": 1940,
    "fps": 24.0,
    "label": ["Explosion"],
    "clips": [[[5.583, 11.903], [11.903, 18.222], [18.222, 24.542]],
                [[36.167, 43.48], [43.48, 50.792]]],
    "clip_captions":[
        [
            "A military tank moving across a barren landscape with low-rise buildings and sparse vegetation. the sky is overcast, and the overall color palette is muted with earthy tones.",
            "A series of images depicting a barren landscape with a few buildings in the background. the foreground consists of a rocky terrain with sparse vegetation. the sky is overcast, and there are no visible people or moving objects.",
            "A silhouette of a person operating a large, mounted weapon on a rocky terrain under a clear sky. the individual appears to be adjusting or aiming the weapon."
        ],
        [
            "A series of images depicting a hazy, dusty environment with buildings in the background. the focus is on a red light source that appears to be moving or flashing, possibly indicating some form of activity or event occurring in the area.",
            "A series of images depicting a military tank moving across a landscape with buildings in the background. the tank is seen from different angles, and there are visible explosions occurring in the distance."
        ]
    ],
    "events": [[5.583, 24.542], [36.167, 50.792]],
    "event_summary": [
    "The anomaly exists, specifically an Explosion anomaly. The basis for judging the anomaly is the presence of an unexpected and sudden release of energy, likely accompanied by a loud noise and visible effects such as smoke, fire, or debris.",
    "The anomaly exists and its specific name is Explosion. The anomaly event is a series of visible explosions occurring in the distance as a military tank moves across a landscape with buildings in the background. The basis for judging the anomaly is the sudden and intense release of energy and light in the form of explosions, which is an unusual and notable event in the otherwise desolate and quiet war-torn environment depicted in the video"],
    "video_summary": "The anomaly exists, specifically named as Explosion. The anomaly event depicts a series of visible explosions occurring in the distance as a military tank moves across a landscape with buildings in the background, characterized by sudden and intense releases of energy and light, accompanied by loud noise and visible effects such as smoke, fire, or debris. The basis for judging the anomaly lies in the unusual and notable nature of these explosions, which stand out against the otherwise desolate and quiet war-torn environment depicted in the video, making them an unexpected and sudden release of energy that grabs attention."
}
\end{minted}
}
\vspace{-4mm}
\caption{
\textbf{An example of hierarchical free-text annotations.}
For each labeled video, the hierarchical free-text annotations include clip-level captions, event-level, and video-level anomaly analysis. Additionally, the temporal boundaries for each event and clip are annotated.
}
\vspace{-5mm}
\label{fig:database2}
\end{figure*}

\noindent\textbf{Annotation Format.}
In Fig.\ref{fig:database2}, we present an example of the hierarchical free-text annotations for a video.

\subsection{Hierarchical Instruction Data Construction}
To construct the instruction dataset, we designed question prompts tailored to different tasks, including \textbf{Caption}, \textbf{Judgment}, \textbf{Description}, and \textbf{Analysis}.
For each instruction item, we randomly selected one prompt from the pool and matched it with the corresponding content from the free-text annotations as the answer.

\noindent\textbf{Caption.}
\begin{enumerate}
\scriptsize{
\item "Describe the video briefly."
\item "Describe the main events that take place in this video."
\item "Give a short description of the video."
\item "What happened in this video?"
\item "Generate a brief caption for the video."
\item "Can you provide a brief description of the video?"
\item "Briefly describe the main subjects and their actions in the video."
\item "Provide a short overview of what happens in the video?"
\item "Describe the key moments that showcase the subjects’ activities in the video."
\item "Describe the sequence of events involving the main subjects in the video."
\item "What activities happen throughout the video?"
\item "Describe the main subjects and their roles in the video."
\item "What key moments stand out in the video?"
\item "What are the primary activities showcased in the video?"
\item "What happens to the main subjects as the video progresses?"
\item "What is a brief overview of what happens in the video?"
\item "Describe the main subjects and their contributions to the video."
\item "Describe the key events in the video."
\item "Describe the video’s main activities."
\item "Can you describe the main action in this video briefly?"
\item "Describe the video clip concisely."
\item "Provide a brief description of the given video clip."
\item "Summarize the visual content of the video clip."
\item "Give a short and clear explanation of the subsequent video clip."
}
\end{enumerate}

\noindent\textbf{Judgement.}
\begin{enumerate}
\scriptsize{
\item "What types of anomalies are shown in the video clip?"
\item "Are there any anomaly events detected in the video?"
\item "Detect and classify the anomaly events in the video."
\item "Identify any abnormal behaviors depicted in the video."
\item "Determine whether there are anomaly events in the video and the specific name of the anomaly."
\item "What anomalies can be identified in the video?"
\item "What categories of anomalies can be found in the video?"
\item "Could you point out any abnormal actions in the video?"
\item "Point out the abnormal actions in the video."
\item "Are there anomalies observed in the video clip?"
}
\end{enumerate}

\noindent\textbf{Description.}
\begin{enumerate}
\scriptsize{
\item "Describe the anomaly events observed in the video."
\item "Could you describe the anomaly events observed in the video?"
\item "Could you specify the anomaly events present in the video?"
\item "Give a description of the detected anomaly events in this video."
\item "Could you give a description of the anomaly events in the video?"
\item "Provide a summary of the anomaly events in the video."
\item "Could you provide a summary of the anomaly events in this video?""
\item "What details can you provide about the anomaly in the video?"
\item "How would you detail the anomaly events found in the video?"
\item "How would you describe the particular anomaly events in the video?"
}
\end{enumerate}

\noindent\textbf{Analysis.}
\begin{enumerate}
\scriptsize{
\item "Why do you judge this event to be anomalous?"
\item "Can you provide the reasons for considering it anomalous?"
\item "Can you give the basis for your judgment of this event as an anomaly?"
\item "What led you to classify this event as an anomaly?"
\item "Could you provide the reasons for considering this event as abnormal?"
\item "What evidence do you have to support your judgment of this event as an anomaly?"
\item "Can you analyze the factors contributing to this anomalous event?"
\item "Could you share your analysis of the anomalous event?"
\item "What patterns did you observe that contributed to your conclusion about this event being an anomaly?"
\item "How do the characteristics of this event support its classification as an anomaly?"
}
\end{enumerate}

\subsection{Data Samples.}
\begin{figure}[h!]
\centering
\includegraphics[width=0.48\textwidth]{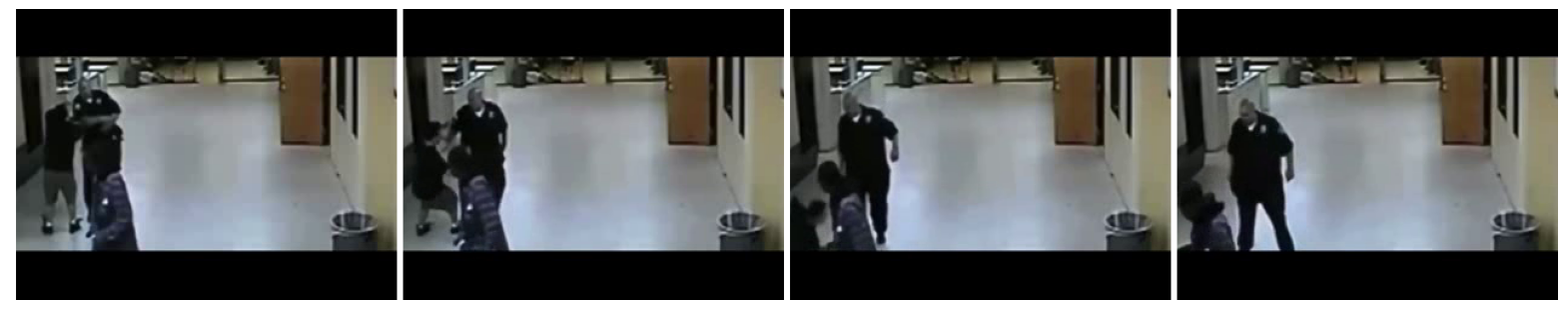}
\vspace{-6mm}
\scriptsize{
\begin{minted}[linenos, frame=lines, breaklines]{json}
{
    "id": "Abuse007_x264_E0C2",
    "type": "clip",
    "video": "ucf-crime/clips/train/Abuse007_x264_E0C2.mp4",
    "conversations": [
    {"from": "human",
    "value": "Describe the key events in the video."},
    {"from": "gpt",
    "value": "The man in shorts was slapped to the ground by the strong man opposite."}
    ]
}
\end{minted}
}
\vspace{-4mm}
\caption{
\textbf{An example of clip-level instruction data.}
}
\label{fig:ins_clip}
\end{figure}

\begin{figure}[h!]
\centering
\includegraphics[width=0.48\textwidth]{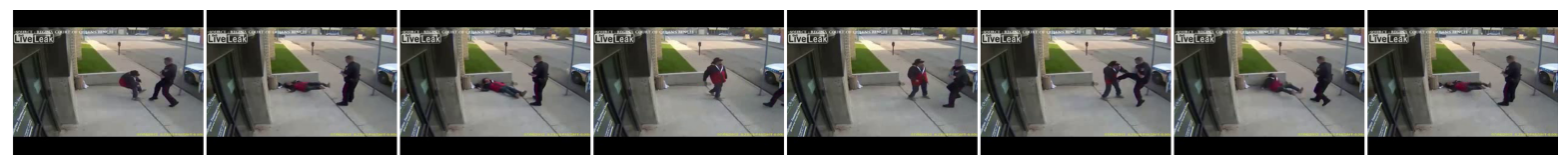}
\vspace{-8mm}
\scriptsize{
\begin{minted}[linenos, frame=lines, breaklines]{json}
{
    "id": "Abuse018_x264_E0",
    "type": "event",
    "video": "ucf-crime/events/train/Abuse018_x264_E0.mp4",
    "conversations": [
    {"from": "human",
    "value": "Identify any abnormal behaviors depicted in the video."},
    {"from": "gpt",
    "value": "An anomaly exists, specifically an instance of Police Brutality."},
    {"from": "human",
    "value": "Could you provide a summary of the anomaly events in this video?"},
    {"from": "gpt",
    "value": "The anomaly event involves the police officer kicking the man who rushed towards him, causing the man to fall heavily to the ground with his head on his back."},
    {"from": "human",
    "value": "Could you share your analysis of the anomalous event?"},
    {"from": "gpt",
    "value": "The excessive use of force by the police officer, which deviates from standard protocol and may be considered an abuse of authority."}
    ]
}
\end{minted}
}
\vspace{-4mm}
\caption{
\textbf{An example of event-level instruction data.}
}
\label{fig:ins_event}
\end{figure}

\begin{figure}[h!]
\centering
\includegraphics[width=0.48\textwidth]{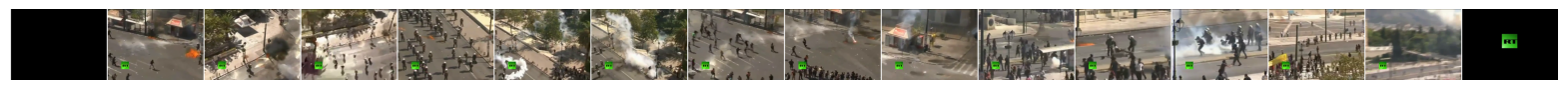}
\vspace{-8mm}
\scriptsize{
\begin{minted}[linenos, frame=lines, breaklines]{json}
{
    "id": "v=xLFbqPzSI2Y__#1_label_B4-0-0",
    "type": "video",
    "video": "xd-violence/videos/train/
    v=xLFbqPzSI2Y__#1_label_B4-0-0.mp4",
    "conversations": [
    {"from": "human",
    "value": "Are there any anomaly events detected in the video?"},
    {"from": "gpt",
    "value": "The anomaly exists and its specific name is the Riot anomaly."},
    {"from": "human",
    "value": "Could you give a description of the anomaly events in the video?"},
    {"from": "gpt",
    "value": "The anomaly event depicts a chaotic and intense urban unrest scene, starting with a chaotic street scene where individuals are running away from an area of fire and smoke, with people appearing to be in a state of panic or urgency, and some being sprayed with water by individuals in uniform. The scene escalates to show a growing fire in the background, police officers in riot gear standing on the street, and a formation of police officers spread out across the road, suggesting an ongoing riot or civil unrest. "},
    {"from": "human",
    "value": "Can you give the basis for your judgment?"},
    {"from": "gpt",
    "value": "The presence of multiple indicators of a riot or emergency situation, including fire, smoke, and people fleeing in panic, as well as the sudden appearance of flames, smoke, and police in riot gear, and the presence of violent and destructive events, such as fires and explosions, and the response of authorities, including the use of tear gas or pepper spray, which deviate significantly from normal, peaceful urban activity."}
    ]
}
\end{minted}
}
\vspace{-4mm}
\caption{
\textbf{An example of video-level instruction data.}
}
\vspace{2mm}
\label{fig:ins_video}
\end{figure}

To facilitate understanding, we provide the final constructed instruction data at various temporal granularities, including \textbf{clip-level}, \textbf{event-level}, and \textbf{video-level}, as shown in Fig.~\ref{fig:ins_clip}, Fig.~\ref{fig:ins_event}, and Fig.~\ref{fig:ins_video}.

\vspace{-4mm}
\vspace{4mm}

\begin{figure}[h!]
\centering
  \vspace{-4mm}
  \includegraphics[width=0.3\textwidth]{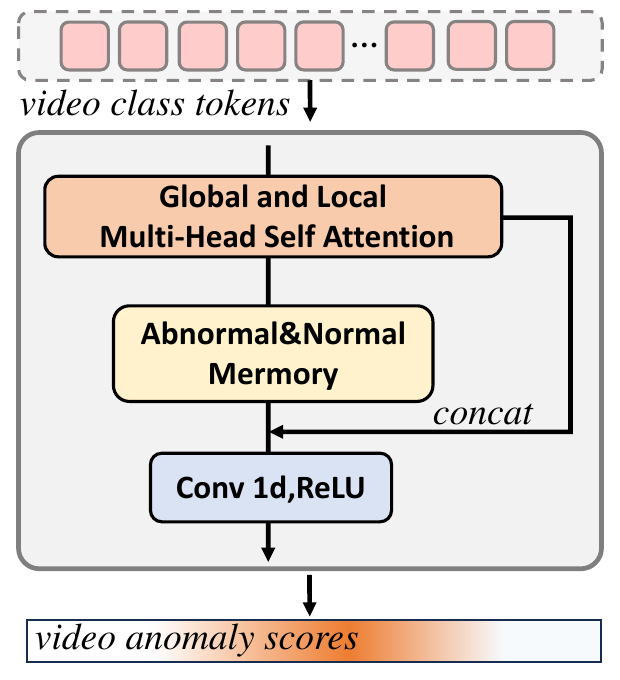}
  \caption{
  \textbf{Architecture of the Anomaly Scorer (UR-DMU~\cite{URDMU}).}
  }
  \vspace{-4mm}
\label{baseline_methods}
\end{figure}

\begin{figure*}[ht!]
\centering\setlength{\abovecaptionskip}{0.1cm}
\includegraphics[width=\textwidth]{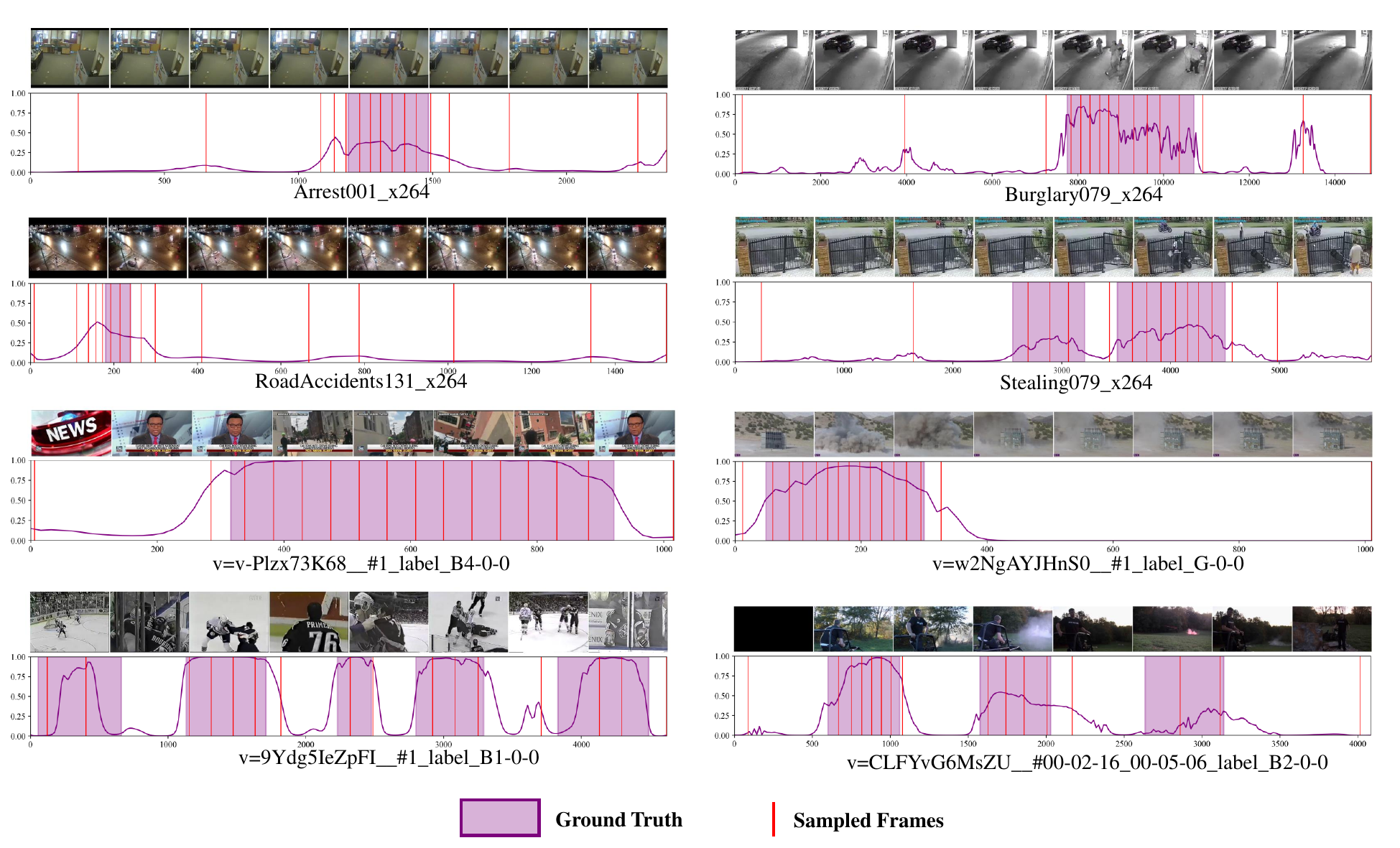}
\caption{\textbf{Visualization results of anomaly scores and sampled frames output by the Anomaly-focused Temporal Sampler.}.}
\vspace{-2mm}
\label{fig:temporal_sampler}
\end{figure*}

\section{Details of the Anomaly Scorer}
\subsection{Model Architecture}
We use UR-DMU~\cite{URDMU} as the anomaly scorer in our Anomaly-focused Temporal Sampler.
As shown in Fig.~\ref{baseline_methods}, UR-DMU utilizes a Global and Local Multi-Head Self Attention (GL-MHSA) module to capture both long-range and short-range temporal relationships among video snippets. Furthermore, UR-DMU introduces two memory banks to store and differentiate abnormal and normal prototypes, thereby maximizing the margins between these two representations.
In order to learn discriminative representations, UR-DMU uses triplet loss to increase the feature distance after interacting with different memories. Simultaneously, it utilizes KL loss to constrain the normal memory to follow a Gaussian distribution, accounting for the variance introduced by noise.
Furthermore, We leveraged the event-level anomaly boundaries obtained during the annotation phase to generate frame-level labels and computed a binary cross-entropy loss, \ie, ${\mathcal L}_{AS}$, which is simple yet effective.
Thus, the loss function for the anomaly scorer is defined as follows:
\begin{equation}
{\mathcal L} = {\mathcal L}_{AS} + {\mathcal L}_{triplet} + {\mathcal L}_{kl}
\end{equation}

\begin{table*}[t!]
\centering
\caption{
\textbf{Comparision with related multimodal/explanable VAU methods and benchmarks.}
HIVAU-70k provides accurate temporal annotations and hierarchical anomaly-related free-text annotations.}
\setlength{\tabcolsep}{4pt}
\resizebox{0.9\textwidth}{!}{
\begin{tabular}{cccccccc}
\toprule
\multicolumn{1}{c}{\multirow{2}{*}{\textbf{Methods}}}  & \multirow{2}{*}{\textbf{\#Catogories}} & \multirow{2}{*}{\textbf{\#Samples}} &\multicolumn{3}{c}{\textbf{Text }} &\multirow{2}{*}{\textbf{Temp. Anno.}} &\multirow{2}{*}{\textbf{MLLM tuning }} \\
\cmidrule{4-6} 
\multicolumn{1}{c}{} & & & \multicolumn{1}{c}{clip-level} & \multicolumn{1}{c}{event-level} & \multicolumn{1}{c}{video-level}  & \\
\midrule
UCA~\citep{yuan2023surveillance} & 13 & 23,542 & \True & \False & \False  & \True & \False \\
LAVAD~\citep{zanella2024harnessing} & N/A & N/A & \True & \False & \True & \False & \False\\
VAD-VideoLLama~\citep{lv2024video} & 13/7 & 2,400 & \False & \False & \True & \False & projection\\
CUVA~\citep{du2024uncovering} &11 &6,000 & \False & \False & \True & \False & \False \\
Hawk~\citep{tang2024hawk} &- & 16,000 & \False & \False & \True & \False & projection \\
\textbf{HIVAU-70k (Ours)} & \textbf{19} & \textbf{70,000} & \True & \True & \True & \True & LoRA\\
\bottomrule
\end{tabular}
}
\label{tab:related_works}
\end{table*}

\subsection{Visualization Results}
In Fig.~\ref{fig:temporal_sampler}, we present visualized results of anomaly scores and sampled frames on the UCF-Crime and XD-Violence test sets. These results demonstrate the accuracy of our method in anomaly detection within complex real-world scenarios, with the sampled frames being concentrated in anomalous regions.

\section{Discussion with related works.}
In Table~\ref{tab:related_works}, we provide a comprehensive comparison with related works in terms of benchmarks and methods.

\noindent\textbf{Summary of related works}: Recently, there has been substantial research on multi-modal Video Anomaly Understanding, making significant contributions to advancing open-world anomaly understanding. LAVAD~\cite{zanella2024harnessing} utilized several pre-trained foundational models to offer a training-free explainable VAD process. VAD-VideoLLaMA~\cite{lv2024video}, designed a three-phase training method to finetune Video-LLaMA in the VAD domain. CUVA~\cite{du2024uncovering} introduced a dataset and metric for evaluating causation understanding of video anomalies. Hawk~\cite{tang2024hawk} constructed an instruction dataset and finetuned a video-language framework that incorporates both motion and video information.

\noindent\textbf{Difference and Advantages of our proposed benchmark and method}:
\begin{itemize}
    \item We develop a semi-automated annotation engine that scales hierarchical anomaly annotation efficiently, combining manual refinement with LLM-based annotation to maintain high-quality data across multiple granularities, resulting in over \textbf{70,000} annotations at clip, event, and video levels, which significantly surpasses previous datasets in scale.
    \item UCA~\cite{yuan2024towards} only provides clip-level captions, overlooking the understanding of anomalies across longer time spans. CUVA~\cite{du2024uncovering} and Hawk~\cite{tang2024hawk}, on the other hand, only offer video-level instruction data, neglecting finer-grained visual perception and anomaly analysis. In contrast, our proposed HIVAU-70k takes a multi-temporal granularity perspective, offering more comprehensive and diverse anomaly annotations for open-world anomaly detection. It enables progressive and comprehensive learning, from short-term visual perception to long-term anomaly reasoning.
    \item We propose the \textbf{Anomaly-focused Temporal Sampler (ATS)}, integrated with a multi-modal visual-language model. Benefiting from the precise temporal annotations we provide, the ATS is able to focus on anomaly-dense video segments. This integration significantly improves efficiency and accuracy in long-video anomaly detection.
\end{itemize}

\section{More Qualitative Results.}
\begin{figure*}[t]
\centering\setlength{\abovecaptionskip}{0.1cm}
\includegraphics[width=\textwidth]{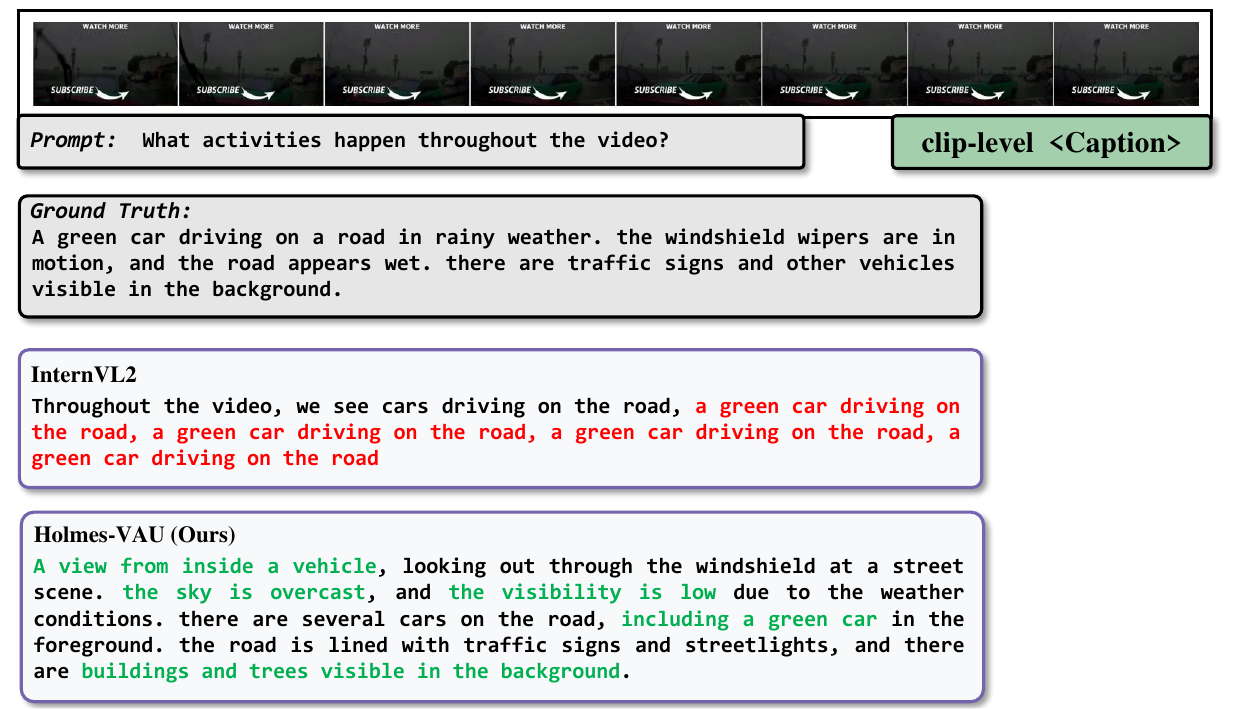}
\includegraphics[width=\textwidth]{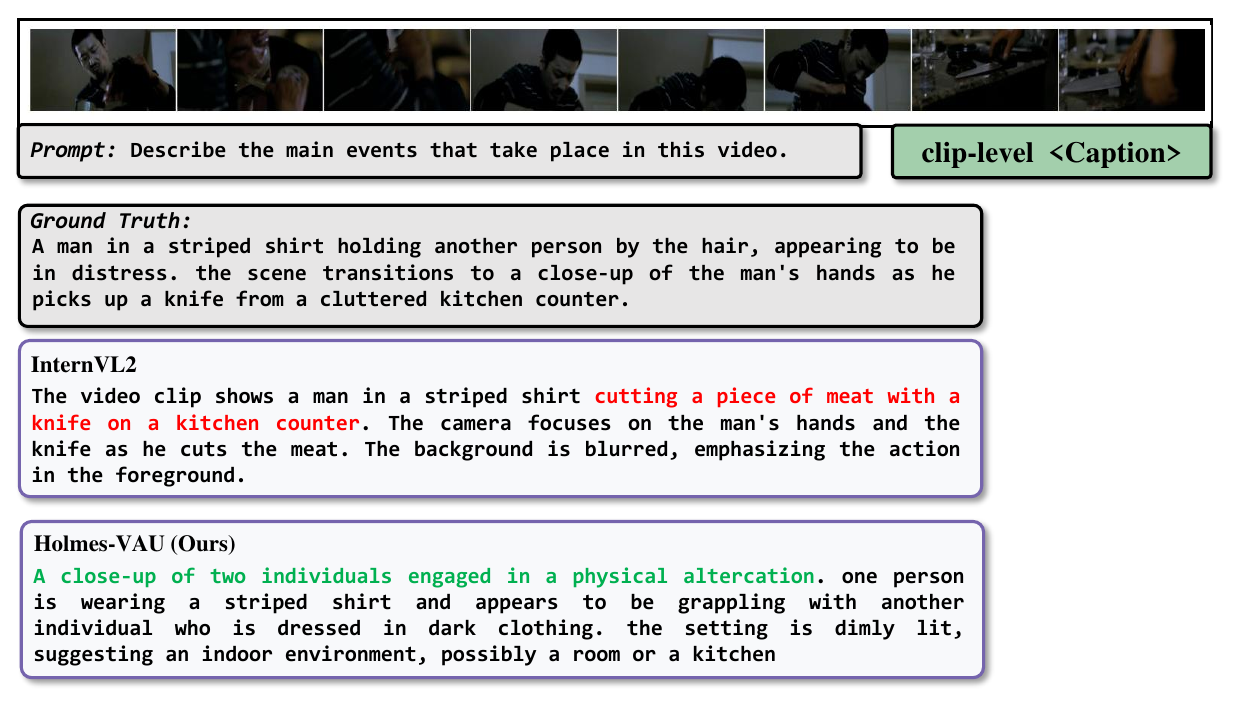}
\caption{\textbf{Qualitative comparison of anomaly understanding explanation with our baseline model, \ie, InternVL-2B.}
Correct and wrong explanations are highlighted in \textcolor[rgb]{0,0.6875,0.3125}{green} and \textcolor{red}{red}, respectively.
}
\vspace{-4mm}
\label{fig:qualitative1}
\end{figure*}

\begin{figure*}[h]
\centering\setlength{\abovecaptionskip}{0.1cm}
\includegraphics[width=\textwidth]{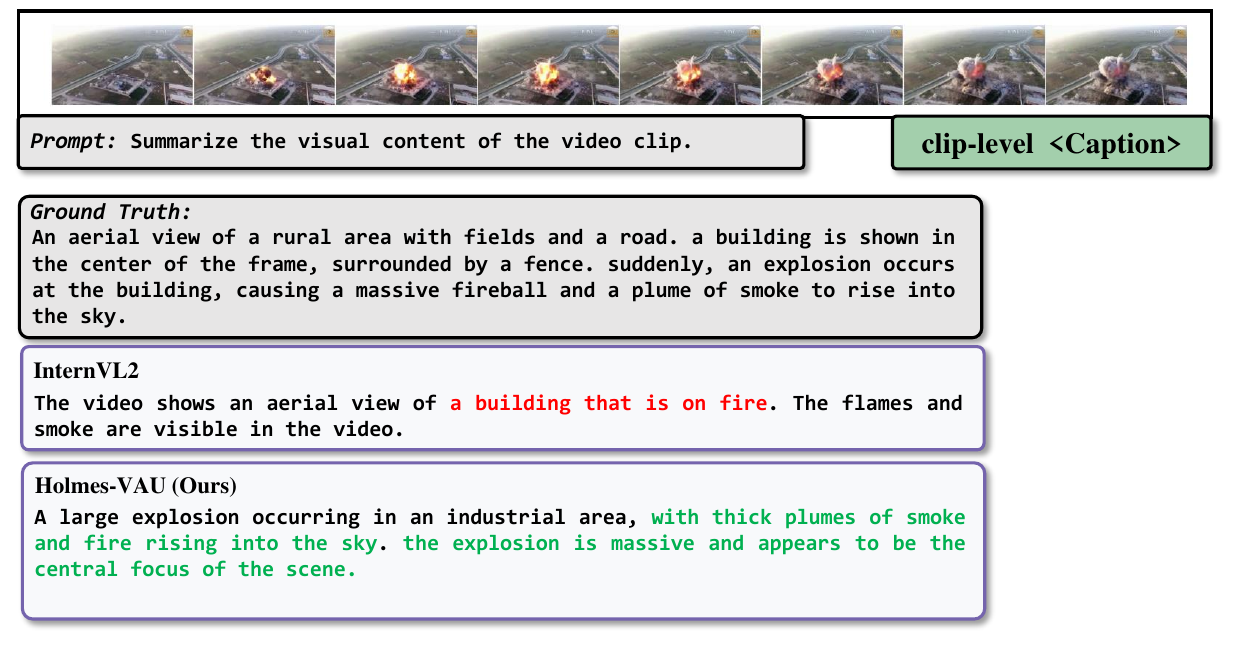}
\includegraphics[width=\textwidth]{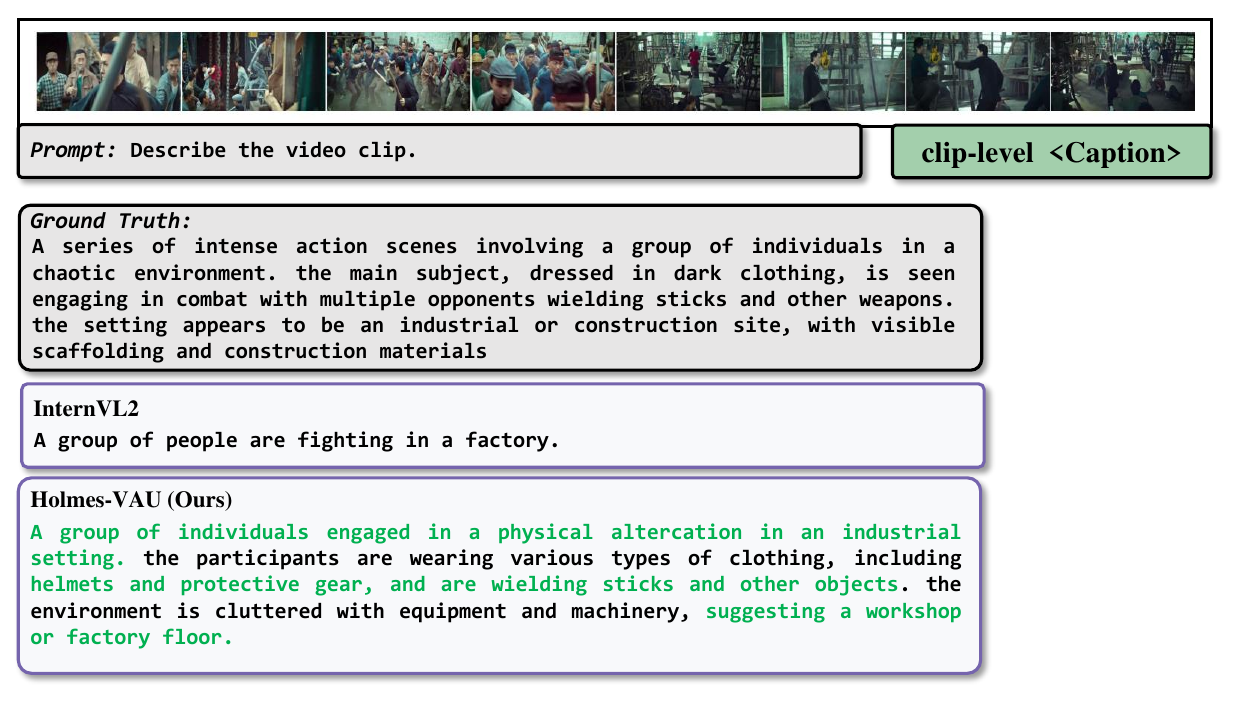}
\caption{\textbf{Qualitative comparison of anomaly understanding explanation with our baseline model, \ie, InternVL-2B.}
Correct and wrong explanations are highlighted in \textcolor[rgb]{0,0.6875,0.3125}{green} and \textcolor{red}{red}, respectively.
}
\vspace{-4mm}
\label{fig:qualitative2}
\end{figure*}

\begin{figure*}[h]
\centering\setlength{\abovecaptionskip}{0.1cm}
\includegraphics[width=\textwidth]{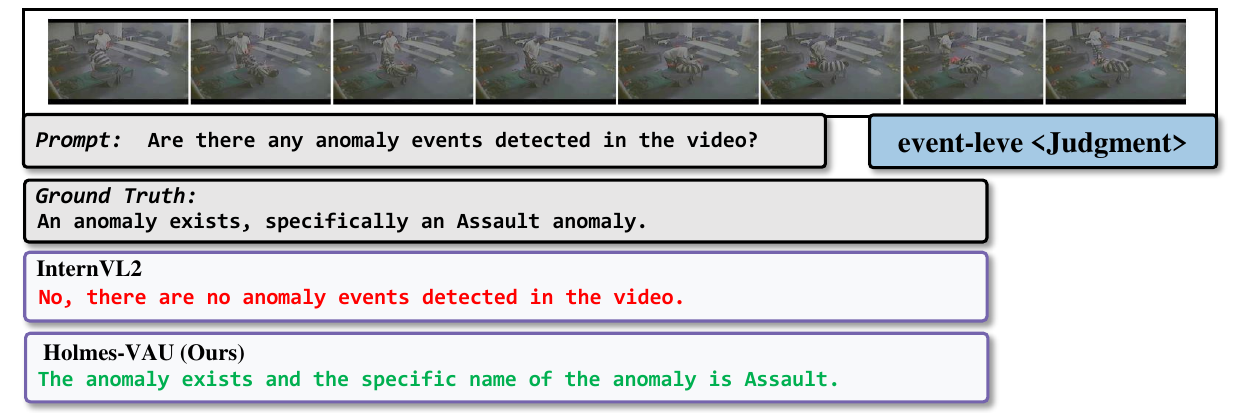}
\includegraphics[width=\textwidth]{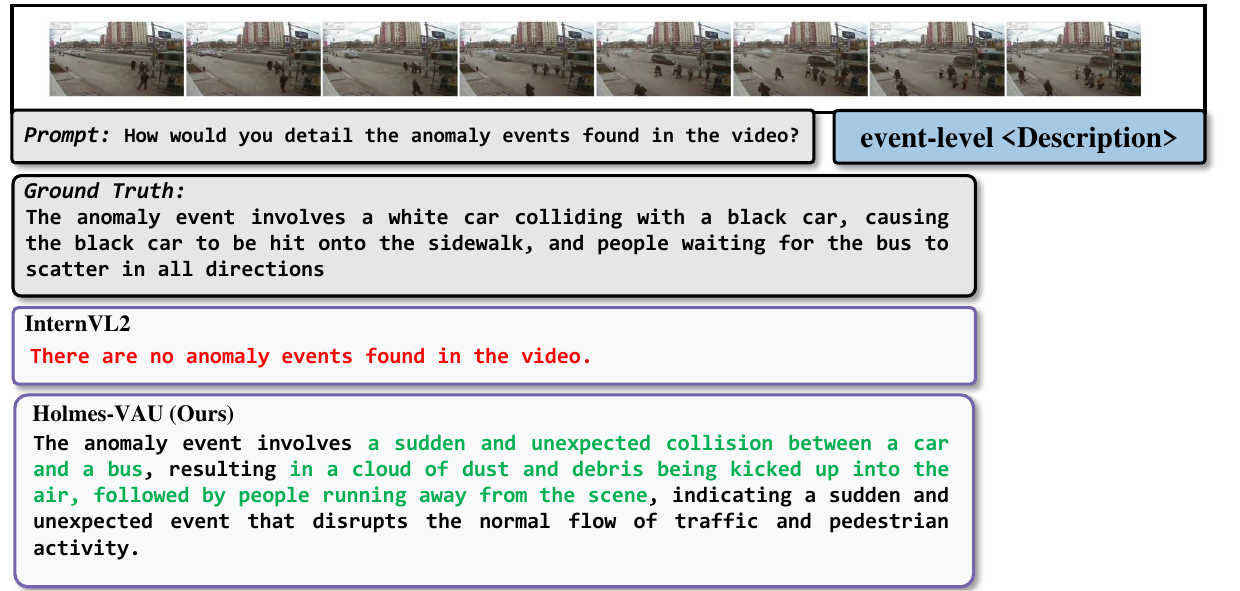}
\includegraphics[width=\textwidth]{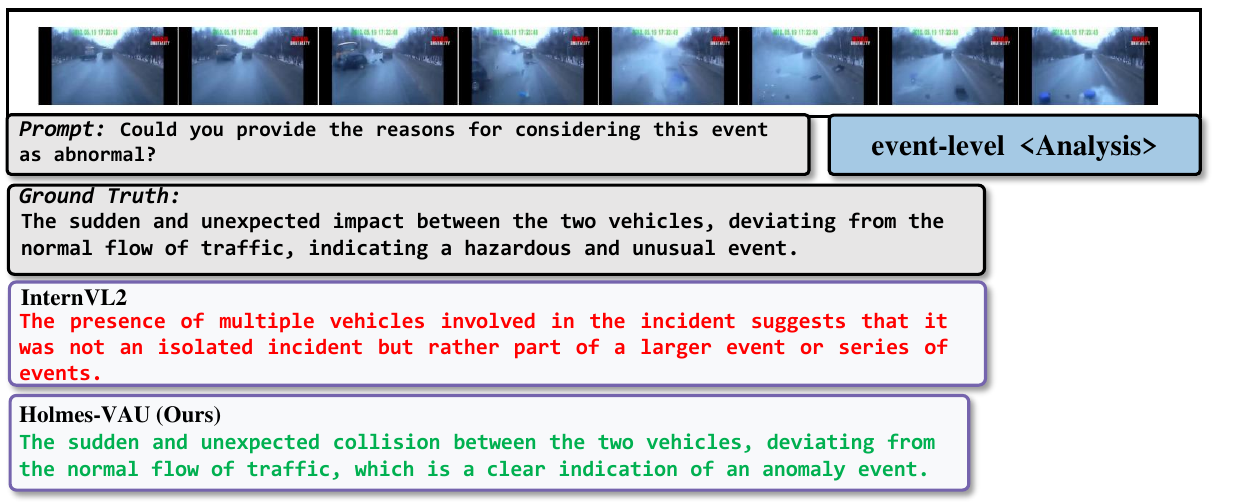}
\caption{\textbf{Qualitative comparison of anomaly understanding explanation with our baseline model, \ie, InternVL-2B.}
Correct and wrong explanations are highlighted in \textcolor[rgb]{0,0.6875,0.3125}{green} and \textcolor{red}{red}, respectively.
}
\vspace{-4mm}
\label{fig:qualitative3}
\end{figure*}

\begin{figure*}[h]
\centering\setlength{\abovecaptionskip}{0.1cm}
\includegraphics[width=\textwidth]{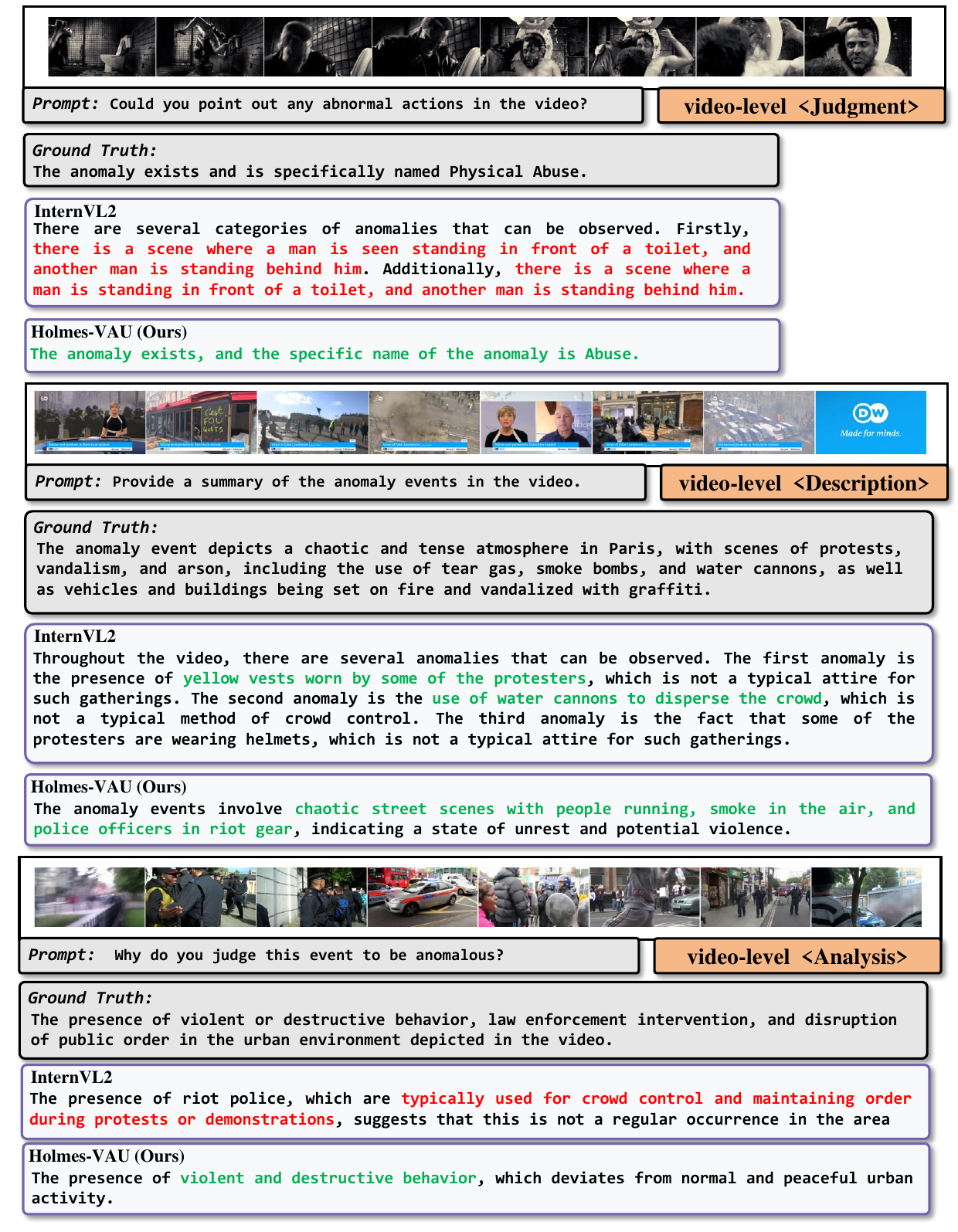}
\caption{\textbf{Qualitative comparison of anomaly understanding explanation with our baseline model, \ie, InternVL-2B.}
Correct and wrong explanations are highlighted in \textcolor[rgb]{0,0.6875,0.3125}{green} and \textcolor{red}{red}, respectively.
}
\vspace{-4mm}
\label{fig:qualitative4}
\end{figure*}

As shown from Fig.~\ref{fig:qualitative1} to Fig.~\ref{fig:qualitative4}, we present the output of explainable text generated by Holmes-VAU compared with the base model, \ie, InternVL-2B~\cite{chen2024far}. The results demonstrate significant improvements in the model's visual perception and anomaly analysis capabilities after fine-tuning on HIVAU-70k.

\section{Limitations and Future Work.}
While our work demonstrates significant strides in multi-granular video anomaly understanding, several limitations present opportunities for future enhancement. First, optimizing for real-time streaming remains a challenge. Our sparse sampling approach improves efficiency, but further refinement is necessary for seamless deployment in streaming contexts. Additionally, our work has so far focused on surveillance data, extending our framework to other domains, such as industrial monitoring and medical diagnostics, will help validate its generalization capabilities. Lastly, integrating additional sensory data, like audio, and scalable hierarchical annotation could enhance anomaly detection and broaden applicability.

\end{document}